\documentclass[11pt]{article}

\usepackage[preprint]{acl}

\usepackage{times}
\usepackage{latexsym}
\usepackage[T1]{fontenc}
\usepackage[utf8]{inputenc}
\usepackage{microtype}
\usepackage{inconsolata}
\usepackage{graphicx}

\usepackage{enumitem}
\usepackage{makecell}
\usepackage{booktabs}
\usepackage[table]{xcolor}
\usepackage{amsmath}
\usepackage{amssymb}
\usepackage{multirow}
\usepackage{pifont}
\usepackage{tabularx}
\newcolumntype{Y}{>{\raggedright\arraybackslash}X}
\usepackage{algorithm}
\usepackage{algorithmic}
\usepackage[normalem]{ulem}

\usepackage{tcolorbox}
\tcbuselibrary{breakable}
\newtcolorbox{promptbox}[1][]{
  colback=gray!4, colframe=gray!55, boxrule=0.5pt, arc=2pt,
  left=6pt, right=6pt, top=5pt, bottom=5pt,
  colbacktitle=gray!12, coltitle=black,
  fonttitle=\small\bfseries, breakable, title=#1}

\title{LLMs Silently Correct African American English: Auditing and Mitigating Dialect Bias via Activation Steering}

\author{
  Huan Wu$^{1,2,3}$ \quad Ali Emami$^{4}$ \quad Muhammad Furquan Hassan$^{1}$ \quad Osaretin Igbinoba$^{5, \#}$ \\ \textbf{Osakpolor Idusuyi}$^{6,\#}$\quad  \textbf{Osamede Igbinoba}$^{7,\#}$ \quad  \textbf{Faiza Khan Khattak}$^{8}$ \\ \textbf{Laleh Seyyed-Kalantari}$^{1,2,3,9}$\thanks{Corresponding author. \quad \# Equal contribution.} \\
  $^{1}$York University \quad $^{2}$Vector Institute \quad $^{3}$ Connected Minds \quad $^{4}$Emory University \\\quad $^{5}$Wilfrid Laurier University \quad $^{6}$University of Toronto \quad $^{7}$University of Guelph \quad \\
  $^{8}$Monark Health \quad $^{9}$CIFAR\\
  \texttt{$^{*}$lsk@yorku.ca}
}

\begin{document}
\maketitle

\begin{abstract}
African American English (AAE) dialect, spoken by over 30 million people, is routinely misinterpreted and ``corrected'' by large language models (LLMs). Across six instruction-tuned open-weight LLMs (14B to 70B), we show dialect bias, a systematic preference for Standard American English (SAE) continuations even when the preceding context is in AAE. Here, we present a framework for auditing and mitigating dialect bias. For auditing, we introduce conditional Dialect Group Invariance (cDGI), which reduces sensitivity to translation drift by restricting
evaluation to prediction-stable, annotator-natural examples, and a feature-level localization analysis that identifies which AAE markers most strongly trigger bias. We find that syntactic properties, especially negative concord (e.g., ``ain't nobody''), are the strongest bias triggers. For mitigation, we apply test-time activation steering, where causal tracing selects the bias-relevant layers and the model activation is steered using a dialect direction, without training. Steering outperforms prompting, context-aware decoding, and LoRA fine-tuning, with about twice the dialect-matched improvement of the strongest baseline, while largely preserving SAE fluency. We also release \textsc{Real-AAE}, the largest real-AAE parallel corpus to date: 17{,}479 AAE/SAE/AAE\textsubscript{back} triplets from natural tweets (3.5 $\times$ larger than the largest prior real-AAE parallel resource), validated automatically (BERTScore F1 = 0.95) and by three native AAE speakers (91.3\% semantic agreement). Our corpus and code are available at \url{https://anonymous.4open.science/r/dialect}.

\end{abstract}

\section{Introduction}

\begin{figure}[t]
    \centering
    \includegraphics[width=\linewidth]{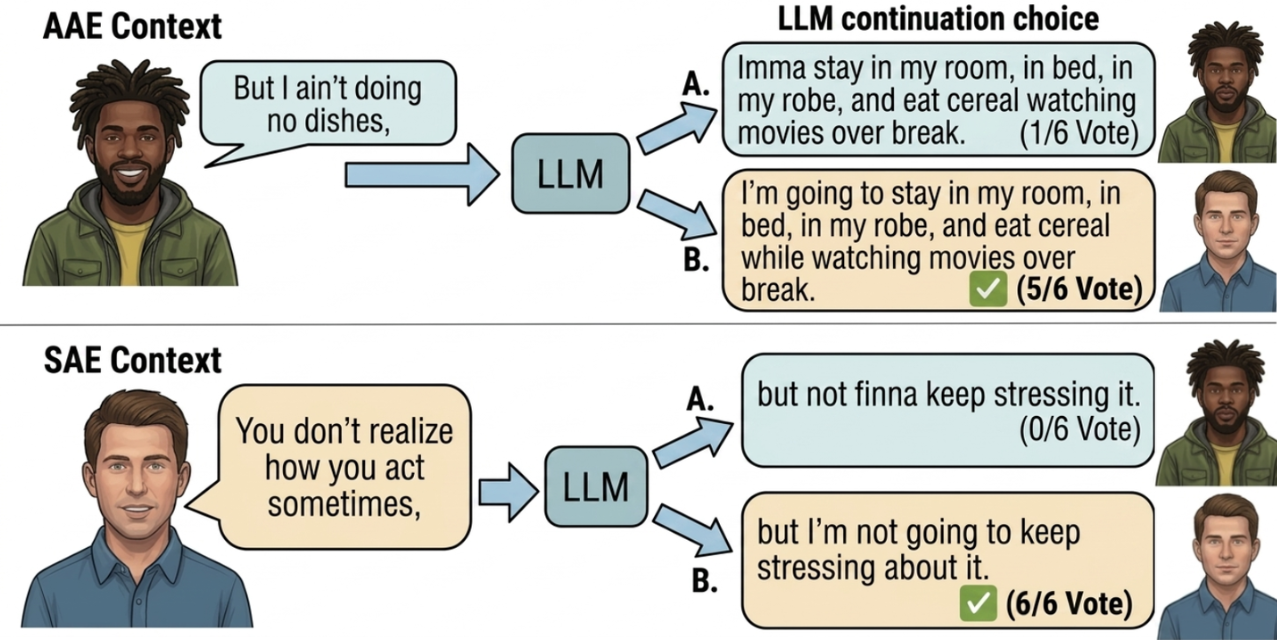}
    \caption{Dialect preference bias on a real AAE tweet. In a multiple-choice continuation task, five out of six LLMs prefer the SAE continuation over a fluent AAE alternative under an AAE context, and unanimously prefer SAE under an SAE context. 
    }
    \label{fig:teaser}
\end{figure}

Large language models (LLMs) silently
``correct'' African American English (AAE). Given the AAE context
\textit{``But I ain't doing no dishes''}, five out of six LLMs we
tested prefer the Standard American English (SAE) continuation
\textit{``I'm going to stay in my room''} over the valid
AAE alternative \textit{``Imma stay in my room''}
(Fig.~\ref{fig:teaser}).

We call this behavior \textit{dialect preference bias}: the systematic tendency of LLMs to favor SAE over valid varieties such as AAE, even when the user is writing in AAE. Treating AAE as something to be corrected is  a form of linguistic discrimination \citep{blodgett2021sociolinguistically}, since AAE is a rule-governed variety
with its own consistent grammar and phonology, spoken by over 30
million people \citep{lin-etal-2025-assessing}.

This preference matters most when LLMs continue or edit user
text, since it pushes the text toward SAE and replaces the user's
own voice. Black users find such corrections hurtful
\citep{basoah2025should} and want to choose when an LLM uses AAE
\citep{sandoval-etal-2025-llm}. The preference also appears in
reward models \citep{mire2025rejecteddialectsbiasesafrican} and in
sentiment labels, which differ between AAE and SAE versions of the
same text \citep{Hassan_Khattak_Seyyed-Kalantari_2025} for 15\% to
53\% of pairs in our audit (\S\ref{sec:sentiment-results}). As LLMs
move into hiring, healthcare, and content moderation, this can
disadvantage AAE speakers \citep{hofmann2024dialect}.

Currently, work on AAE dialect bias has three key limitations
(Table~\ref{tab:comparison} in Appendix~\ref{app:prior-work}). First, most benchmarks generate AAE
\emph{synthetically} from SAE via rule-based or LLM-driven
translation \citep{ziems2022value, ziems-etal-2023-multi,
gupta2024aavenue, gupta2025endive}, producing AAE that
underestimates real-world dialect effects
\citep{lin-etal-2025-assessing, gupta2025endive}. Second, real-AAE
studies are narrow in scope. They evaluate either classification
consistency \citep{Hassan_Khattak_Seyyed-Kalantari_2025},
generated text \citep{groenwold2020, deas-etal-2023-evaluation,
fleisig2024linguisticbiaschatgptlanguage}, or reward-model
scoring \citep{mire2025rejecteddialectsbiasesafrican}, but none
combine discriminative and generative auditing, identify which
linguistic features trigger bias, or offer mitigation. Third, the
few existing mitigation methods require retraining
\citep{ziems-etal-2023-multi, dacon-etal-2022-evaluating}, architectural changes
\citep{srirag2025predictingtargetwordgameplaying,
zhou2025dialectgenbenchmarkingimprovingdialect}, or auxiliary
translation pipelines that erase the dialect altogether
\citep{klisura2025multiagentframeworkmitigatingdialect}.

Starting from a corpus of tweets by native AAE speakers
\citep{blodgett-etal-2016-demographic}, we construct
\textsc{Real-AAE}, a corpus of 17{,}479
AAE/SAE/AAE\textsubscript{back} triplets via translation and
back-translation, validated both automatically and by three
native AAE speakers. This AAE-first
construction preserves dialect features that synthetic AAE misses.

We then audit six instruction-tuned open-weight LLMs (14B to 70B) along two
axes: \textit{discriminative consistency}, via our new conditional
Dialect Group Invariance (cDGI) metric, which isolates model bias
from translation drift; and \textit{generative preference}, via
perplexity, log-probability, and multiple-choice continuation. A
feature-level analysis further identifies which AAE markers
trigger the bias.

Finally, we mitigate the bias with activation steering, which has
not previously been applied to dialect bias against AAE. Our method
combines activation steering \citep{turner2023activation,
rimsky-etal-2024-steering} with causal tracing
\citep{meng2022locating}. Causal tracing selects the layers that
carry the dialect preference and sets how strongly each one is
steered, and an extracted dialect direction is then added at
inference without any training. Concurrent work steers the
generation of Arabic dialects \citep{elozeiri2026dialects}, but
does not study dialect bias. Our analysis yields three main
findings: (i) all six models prefer SAE continuations even under
AAE context, defaulting to ``correcting'' the dialect; (ii)
syntactic properties are the strongest bias triggers across all
tested models; and (iii) steering shifts dialect preference more
than prompting, context-aware decoding, and LoRA fine-tuning,
while largely preserving SAE fluency.

\paragraph{Contributions.}
\begin{itemize}[itemsep=1pt, topsep=0pt, parsep=0pt, leftmargin=*]
\item We release \textsc{Real-AAE}, the largest real-AAE
parallel corpus to date: 17{,}479
AAE/SAE/AAE\textsubscript{back} triplets from natural tweets, 3.5
$\times$ larger than the largest prior real-AAE parallel resource, validated both
automatically and by three native AAE speakers.

\item We audit six LLMs with a framework that combines cDGI, which
reduces sensitivity to translation drift, with
generative-preference metrics and a feature-level localization
analysis.

\item We propose, to our knowledge, the first activation-steering
method for mitigating bias against AAE. It requires no training
and, on all six models, improves dialect-matched behavior more
than prompting, CAD, and LoRA fine-tuning while largely preserving
SAE fluency.
\end{itemize}

\begin{figure*}[t]
    \centering
    \includegraphics[width=1.0\linewidth]{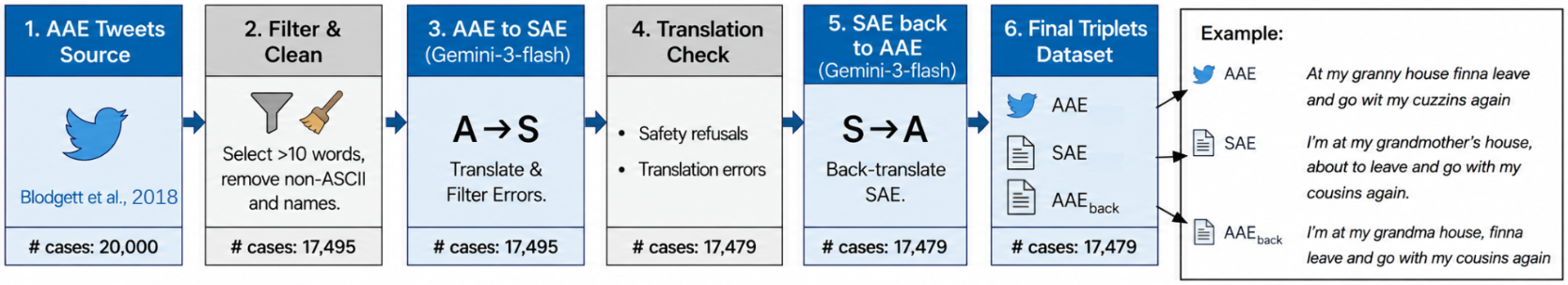}
    \caption{\textsc{Real-AAE} construction pipeline. Starting from authentic AAE tweets, we filter the data, translate AAE into SAE, remove invalid outputs, and back-translate the SAE sentences into AAE to build the final triplet dataset.
    }
    \label{fig:Pipeline-aae-sae}
\end{figure*}

\section{Related Work}
\label{sec:related-work}

\noindent\textbf{Dialect Bias in NLP.} Dialect bias has been
documented across natural language processing tasks, including
language identification \citep{blodgett2021sociolinguistically},
part-of-speech tagging \citep{dacon-2022-towards},
natural language inference \citep{dacon-etal-2022-evaluating},
hate-speech and toxicity detection \citep{sap-etal-2019-risk,
davidson2019, sap2022}, automatic speech recognition
\citep{koenecke2020racial}, text generation
\citep{groenwold2020, deas-etal-2023-evaluation}, and downstream decision-making
about employability, criminality, and personality
\citep{hofmann2024dialect}. Most notably,
\citet{hofmann2024dialect} show that dialect prejudice persists
in instruction-tuned LLMs even after overt racial biases are
reduced, motivating dialect-specific audits of models.
\citet{blodgett2021sociolinguistically} argues that such work
should be grounded in how language varieties are used and
perceived, which motivates our use of naturally occurring AAE and
native-speaker validation.

\noindent \textbf{AAE Bias Benchmarks.} AAE bias benchmarks differ in how they obtain AAE text. \textit{Synthetic}
benchmarks transform SAE into AAE via rule-based or LLM-driven
translation \citep{ziems2022value, ziems-etal-2023-multi,
gupta2024aavenue, gupta2025endive}, an approach shown to
underestimate real-world dialect effects
\citep{lin-etal-2025-assessing, gupta2025endive}.
\textit{Real-AAE} resources instead draw text directly from AAE
speakers \citep{blodgett2021sociolinguistically, groenwold2020,
deas-etal-2023-evaluation,
Hassan_Khattak_Seyyed-Kalantari_2025,
fleisig2024linguisticbiaschatgptlanguage,
mire2025rejecteddialectsbiasesafrican}, but each 
targets a single evaluation mode: classification consistency
\citep{Hassan_Khattak_Seyyed-Kalantari_2025}, the quality of text
generated from AAE and SAE prompts \citep{groenwold2020,
deas-etal-2023-evaluation}, generative responses
\citep{fleisig2024linguisticbiaschatgptlanguage}, or reward-model
scoring \citep{mire2025rejecteddialectsbiasesafrican}. These
parallel resources are small: \citet{groenwold2020} pair 2{,}019
AAE tweets with crowd-written SAE versions, which
\citet{hofmann2024dialect} reuse for matched-guise probing, and
\citet{deas-etal-2023-evaluation} provide 346 AAE texts with
counterparts written by AAE speakers. ReDial
\citep{lin-etal-2025-assessing} instead asks AAE speakers to
rewrite about 1{,}200 reasoning prompts, so its AAE is not
naturally occurring. The closest
prior work, \citet{Hassan_Khattak_Seyyed-Kalantari_2025},
introduces Dialect Group Invariance (DGI) on a real-AAE corpus
for sentiment classification. We extend this work with a 3.5$\times$ larger corpus, additional
discriminative and generative metrics, feature-level localization
of bias triggers, and a test-time mitigation
(Table~\ref{tab:comparison}, Appendix~\ref{app:prior-work}).

\noindent \textbf{Dialect Bias Mitigation.} Prior mitigation
approaches fall into three categories. \textit{Training-time}
methods augment training data with dialectal examples
\citep{ziems-etal-2023-multi, dacon-etal-2022-evaluating}, but can degrade SAE accuracy.
\textit{Architectural} methods add dialect-specific modules, such
as the LoRDD adapter
\citep{srirag2025predictingtargetwordgameplaying} or the
encoder-based DialectGen
\citep{zhou2025dialectgenbenchmarkingimprovingdialect}.
\textit{Inference-time} methods translate AAE inputs to SAE before
passing them to the LLM
\citep{klisura2025multiagentframeworkmitigatingdialect}, which
avoids retraining but introduces auxiliary models and erases the
dialect rather than mitigating bias against it. Our approach
instead intervenes directly on the target model's activations at
inference, requiring no retraining, architectural changes,
auxiliary modules, or dialect erasure.

\noindent\textbf{Activation Steering.} Our method builds on activation steering, a mechanistic-interpretability
technique that controls model behavior by adding learned directions to
hidden states at inference. The method has been applied to
general behavioral control \citep{turner2023activation,
zou2023representation, rimsky-etal-2024-steering}, refusal
behavior \citep{arditi2024refusallanguagemodelsmediated}, and
persona traits \citep{chen2025persona}.
\citet{liu2024incontext} extend it to social bias mitigation,
but not to dialect bias. Concurrent work shows that Arabic dialects
can be steered in generation \citep{elozeiri2026dialects}, but it
does not measure or mitigate bias. To our knowledge, ours is the
first application of activation steering to bias against AAE.

\section{Benchmark Construction}
\label{construction}
To audit AAE dialect bias while controlling for translation drift, we construct \textsc{Real-AAE}, an \textit{AAE-first} parallel corpus of (AAE, SAE, $\text{AAE}_{\text{back}}$) triplets built from naturally occurring AAE tweets (Fig.~\ref{fig:Pipeline-aae-sae}).

\paragraph{Source data and filtering.}
We build on the TwitterAAE corpus
\citep{blodgett-etal-2016-demographic}, which identifies tweets from
African American accounts via demographic inference from
geolocation and network analysis. To ensure sufficient content
for sentiment analysis and continuation modeling, we retain
tweets with at least 10 words and strip non-ASCII characters,
user mentions, and URLs, yielding 17{,}495 tweets.

\paragraph{Translation and back-translation.}
We translate each AAE tweet into SAE using gemini-3-flash-preview, then
back-translate the SAE into AAE using the same model (denoted
$\text{AAE}_{\text{back}}$). The back-translation serves two
purposes: (i) it provides a content-preservation check, since
AAE and $\text{AAE}_{\text{back}}$ should remain semantically
aligned if translation did not introduce drift; and (ii) it
enables our cDGI metric (\S\ref{sec:metrics}), which conditions
on translation-stable examples to isolate model bias from
translation drift. Translation failures were rare: of 17,495 attempts, 10 were blocked by the translator's safety filter, 5 were removed for
translator-added explanations, and one returned empty output,
yielding our final \uline{17{,}479 triplets} (error analysis in Appendix~\ref{app:translation-errors}).

\paragraph{Automatic validation.}
For content preservation, we compute BERTScore F1 between AAE
and $\text{AAE}_{\text{back}}$ using Twitter-RoBERTa,\footnote{\raggedright%
\texttt{cardiffnlp/\allowbreak twitter-roberta-base-2022-154m}.} chosen to
match the social-media domain of our inputs. Both mean and median
F1 are \uline{0.95} (standard deviation $0.02$), indicating
strong semantic alignment. For sentiment preservation, we compare
AAE/SAE label agreement using the same Twitter-RoBERTa model;
agreement is highest for positive and negative cases, with most
disagreements concentrated in neutral examples (full results in Appendix~\ref{app:auto-validation}).

\paragraph{Human validation.} 
We further validate a 2{,}000-example subset with three native
AAE speakers from the West Coast and Northeast/Mid-Atlantic, all
of whom self-reported strong familiarity with written AAE.
Annotators judged each example along four dimensions: sentiment
preservation, semantic equivalence, original AAE naturalness,
and $\text{AAE}_{\text{back}}$ naturalness (guidelines in Appendix~\ref{app:guidelines-validation}).
Because several dimensions are dominated by one
label, Fleiss' $\kappa$ \citep{fleiss1971kappa} is deflated by the prevalence paradox, so
we report Gwet's AC1~\citep{gwet2008computing}.
Three-way exact agreement was \uline{90.9\%} (sentiment preservation; $AC1=0.93$),
\uline{91.3\%} (semantic equivalence; $AC1=0.86$), \uline{87.0\%}
(original AAE naturalness; $AC1=0.83$), and \uline{79.6\%}
($\text{AAE}_{\text{back}}$ naturalness; $AC1=0.72$).

\section{Methods}

We define our bias metrics (\S\ref{sec:metrics}), the
feature-level analysis (\S\ref{sec:feature-analysis}), and our
steering method (\S\ref{sec:steering}).

\subsection{Bias Evaluation Metrics}
\label{sec:metrics}

We evaluate dialect bias along two complementary axes:
\textit{discriminative consistency}, via the existing DGI and two-way DGI metrics
\citep{Hassan_Khattak_Seyyed-Kalantari_2025} and our new cDGI
metric; and \textit{generative preference}, via perplexity, a
log-probability bias score, and multiple-choice continuation
computed from model likelihoods.

\subsubsection{Conditional Dialect Group Invariance (cDGI)}

DGI \citep{Hassan_Khattak_Seyyed-Kalantari_2025} measures
AAE--SAE classification agreement, and its two-way variant
additionally requires the same label for $\text{AAE}_{\text{back}}$.
Neither can separate the model's dialect bias from translation
drift: an apparent inconsistency may reflect semantic drift in the
LLM-produced SAE translation or in $\text{AAE}_{\text{back}}$. We therefore
introduce \uline{conditional DGI (cDGI)}, restricted
to examples satisfying two conditions: (i) the predicted label
is preserved under back-translation, controlling for translation drift; and (ii) $\text{AAE}_{\text{back}}$ was judged
natural by our native-AAE annotators, preventing
$\text{AAE}_{\text{back}}$ drift toward SAE from inflating the
conditioning set. Letting $y^{\mathrm{AAE}}, y^{\mathrm{SAE}},
y^{\mathrm{AAE_{back}}}$ denote model predictions on the AAE,
SAE, and back-translated AAE inputs respectively,
$\mathrm{cDGI}$ is defined as
\begin{equation}
\mathrm{cDGI}
=
\Pr\!\left(
y^{\mathrm{AAE}} = y^{\mathrm{SAE}}
\;\middle|\;
y^{\mathrm{AAE}} = y^{\mathrm{AAE_{back}}}
\right),
\end{equation}
computed over examples meeting condition (ii). $\mathrm{cDGI}$
thus isolates the dialect sensitivity from both
translator noise and dialect erasure in the conditioning set.

\subsubsection{Perplexity Analysis}
\label{sec:Perplexity Analysis}

To measure generative preference, we compute conditional
perplexity ($\mathrm{PPL}$) for all four
context$\rightarrow$continuation combinations:
SAE$\rightarrow$SAE, SAE$\rightarrow$AAE, AAE$\rightarrow$SAE,
and AAE$\rightarrow$AAE. Lower $\mathrm{PPL}$ indicates the model
finds a continuation more natural given the context. A
dialect-invariant model should prefer dialect-matched
continuations: lower $\mathrm{PPL}$ for SAE$\rightarrow$SAE than
SAE$\rightarrow$AAE, and lower for AAE$\rightarrow$AAE than
AAE$\rightarrow$SAE.

Our continuation-based experiments
(\S\ref{sec:MC-evaluation},
and \S\ref{sec:log-p}) all require splitting each AAE--SAE pair
into a context and a continuation. We describe the alignment-based segmentation procedure in Appendix~\ref{app:splitting} and show in Appendix~\ref{app:seg-robustness} that the audit findings are robust to the choice of split point.

\subsubsection{Multiple-Choice Evaluation}
\label{sec:MC-evaluation}

We assess explicit preference by presenting each model with a
context (in either SAE or AAE) and two candidate continuations
(one SAE, one AAE), asking which better continues the context.
Option order is randomized to control for positional bias. A
dialect-invariant model should prefer the continuation matching
the context dialect. The prompt template is in Appendix~\ref{app:prompt}.

\subsection{Feature-Level Bias Localization}
\label{sec:feature-analysis}

\subsubsection{Context-specific log-probability bias}
\label{sec:log-p}

For each of the $N$ AAE--SAE pairs, let $x_i^{\mathrm{SAE}},
x_i^{\mathrm{AAE}}$ and $y_i^{\mathrm{SAE}}, y_i^{\mathrm{AAE}}$
denote the matched SAE/AAE contexts and continuations, and let
$P(y \mid x)$ denote the model's likelihood of continuation $y$
given context $x$. We define context-specific SAE and AAE
log-probability ($\mathrm{LP}$) bias scores as
\begin{align}
b^{\mathrm{SAE}}_i
&= \log P(y_i^{\mathrm{SAE}}\mid x_i^{\mathrm{SAE}}) \nonumber\\
&\quad -\log P(y_i^{\mathrm{AAE}} \mid x_i^{\mathrm{SAE}}),
\label{Eq:bSAE} \\
b^{\mathrm{AAE}}_i
&=\log P(y_i^{\mathrm{AAE}}\mid x_i^{\mathrm{AAE}}) \nonumber\\
&\quad -\log P(y_i^{\mathrm{SAE}}\mid x_i^{\mathrm{AAE}}).
\label{Eq:bAAE}
\end{align}
Positive values indicate preference for dialect-matched
continuations; negative values indicate preference for
mismatched ones. The aggregate $\mathrm{LP}$ bias is
\begin{equation}
\mathrm{LP}
=
\frac{1}{N}\sum_{i=1}^{N} \big(b^{\mathrm{SAE}}_i - b^{\mathrm{AAE}}_i\big),
\label{eq:LP}
\end{equation}
where larger positive values indicate overall preference for SAE
across both contexts, and values near zero indicate
dialect-matched behavior.

\subsubsection{Localizing bias by feature category}

To identify which linguistic markers most strongly trigger
dialect preference, we partition AAE sentences by the presence
of each feature category and compare mean $\mathrm{LP}$
(Eq.~\ref{eq:LP}) across categories. Using an AAE grammatical lexicon adapted from
\citet{harris2022exploring}, we group features into four
classes: \textit{Auxiliary} (non-standard auxiliary forms, e.g.,
``we was'', ``finna''), \textit{Aspectual} (habitual and
completive markers, e.g., habitual ``be'' as in ``he be
working''), \textit{Preverbal} (pre-verbal lexical items, e.g.,
``she done finished''), and \textit{Syntactic} (clause-level
constructions, especially negative concord as in ``ain't
nobody''). The full lexicon is in Appendix~\ref{app:lexicon}.

\subsection{Debiasing with Activation Steering}

\label{sec:steering}

If a model treats AAE and SAE inputs differently, their hidden
states should differ along some direction, and adding a vector
along this direction at inference can shift the model's behavior
without retraining. The method has two phases
(Fig.~\ref{fig:bias-mitigation} in the Appendix). \textbf{Phase~1} (one-time,
offline) extracts a per-layer \emph{dialect direction} from
paired AAE--SAE inputs and uses causal tracing to identify which
layers carry the dialect-preference computation. \textbf{Phase~2}
(test-time) injects the dialect direction into those layers,
weighted by their causal importance, with no parameter updates.
Reproducibility details are in Appendix~\ref{reproducibility}.

\paragraph{Phase 1a: Extracting the dialect direction.}
Given $N$ paired inputs
$\{(x_i^{\mathrm{AAE}}, x_i^{\mathrm{SAE}})\}_{i=1}^{N}$, let
$h_l(x)$ denote the layer-$l$ hidden state at the last context
token. We compute the per-example contrast
\begin{equation}
\Delta_{i,l} = h_l(x_i^{\mathrm{AAE}}) - h_l(x_i^{\mathrm{SAE}}),
\label{eq:contrast}
\end{equation}
average across examples, and normalize to obtain a unit-norm
\emph{dialect direction} at each layer $l$:
\begin{equation}
\bar{\Delta}_l = \frac{1}{N}\sum_{i=1}^{N}\Delta_{i,l},
\qquad
\hat{v}_l = \frac{\bar{\Delta}_l}{\lVert \bar{\Delta}_l \rVert_2}.
\label{eq:direction}
\end{equation}
By construction, $\hat{v}_l$ points from SAE-conditioned
activations toward AAE-conditioned activations at layer $l$.

\paragraph{Phase 1b: Identifying causally important layers.}
Not every layer contributes equally to dialect preference. To
isolate the layers where the bias is computed, we use a
corruption-and-restore causal tracing analysis
\citep{meng2022locating}. For each example, we corrupt the input
embeddings at AAE feature-token
positions with Gaussian noise:
\begin{equation}
e_p^{\mathrm{corr}} = e_p + \epsilon_p,
\quad
\epsilon_p \sim \mathcal{N}(0, \sigma^2 I),
\end{equation}
where $\sigma = \alpha \cdot \mathrm{std}(E_{\mathrm{feat}})$,
$E_{\mathrm{feat}}$ is the set of feature-token embeddings in
the example, and $\alpha = 3.0$ is fixed across all models to
produce a clear perturbation signal without overly destabilizing
the representation. We then run the model on the corrupted input
and, at a single site $z = (l, t)$, restore the clean layer-$l$
hidden state at token position $t$. Letting $g^{\mathrm{corr}}$
denote the fully corrupted run and $g_z^{\mathrm{rest}}$ the run
with restoration at site $z$, we measure the effect of
restoration on the AAE-context preference score
(Eq.~\ref{Eq:bAAE}):
\begin{equation}
R_i(z) = b_i^{\mathrm{AAE}}(g^{\mathrm{corr}})
       - b_i^{\mathrm{AAE}}(g_z^{\mathrm{rest}}).
\label{eq:restoration}
\end{equation}
Because $b_i^{\mathrm{AAE}} > 0$ corresponds to the model
preferring the AAE-matched continuation, $R_i(z) < 0$ means
restoration at site $z$ shifts the model \emph{toward}
dialect-matched behavior, while $R_i(z) > 0$ means restoration
amplifies SAE preference. We average over examples and the
relevant token positions $T$ to obtain a per-layer score:
\begin{equation}
R_l = \frac{1}{N \cdot \lvert T \rvert}
\sum_{i=1}^{N} \sum_{t \in T} R_i\big((l, t)\big),
\label{eq:Rl}
\end{equation}
and select the four layers with the most negative $R_l$ values
as the steering set $K$, i.e., the layers whose restoration most
strongly shifts the model toward AAE-matched behavior.

\paragraph{Phase 2: Inference-time steering.}
At inference, we attach forward hooks at each layer $l \in K$
and modify the hidden state at every token position $t$ as
\begin{equation}
h_{l,t}' = h_{l,t} + \beta\, w_l\, \hat{v}_l,
\qquad l \in K,\; \beta \geq 0,
\label{eq:steering}
\end{equation}
where $\beta$ controls the overall steering strength and $w_l$
is a normalized per-layer weight proportional to causal
importance:
\begin{equation}
w_l = \frac{-R_l}{\sum_{j \in K}(-R_j)}, \quad l \in K.
\label{eq:weight}
\end{equation}
Since $\hat{v}_l$ points from SAE-conditioned activations toward
AAE-conditioned activations, adding $\beta w_l \hat{v}_l$
applies stronger steering at layers with larger beneficial
restoration effects. Setting $\beta = 0$ recovers the
unmodified model; larger $\beta$ yields stronger steering.

\section{Experimental Setup}
\label{sec:setup}

\paragraph{Models.}We evaluate six instruction-tuned open-weight LLMs
spanning families and scales (14B to 70B):
Gemma-3-27B \citep{google2025gemma3},
Mistral-Small-3.1-24B \citep{mistralai2025mistralsmall31},
Phi-3-Medium-14B \citep{abdin2024phi3},
Phi-4-14B \citep{abdin2024phi4},
DeepSeek-R1-Distill-Qwen-32B \citep{deepseek2025r1}, and
Llama-3.1-70B \citep{dubey2024llama3}. Hereafter referred to as Gemma-3, Mistral, Phi-3, Phi-4, DeepSeek-R1, and Llama-3.1-70B, respectively.

\paragraph{Dataset.}All evaluation uses \textsc{Real-AAE}
(\S\ref{construction}). For sentiment classification, models
predict positive, negative, or neutral.

\paragraph{Activation steering splits.}For steering experiments,
we use four non-overlapping splits of \textsc{Real-AAE}:
\uline{1{,}200} AAE-feature sentences for causal-tracing layer
selection (such examples provide clearer corruption-and-restore
targets), \uline{7{,}000} AAE--SAE pairs for dialect-direction
extraction, a \uline{1{,}000} pairs validation set, and a \uline{2{,}000}-pair test set. We sweep the steering strength
$\beta \in \{0.1, 0.2, 0.4, 0.6, 0.8, 1.0\}$ on the validation set to study the trade-off between bias reduction and language
modeling ability. With $\beta$ frozen, all reported results are computed once on the test set.

\paragraph{Baselines.} We compare activation steering with the
unsteered model (\textit{Base}), context-aware decoding
(\textit{CAD}; \citealp{shi-etal-2024-trusting}), which up-weights
the context during decoding, and LoRA fine-tuning (\textit{LoRA};
\citealp{hu2022lora}) on dialect-matched continuations from the
direction split. We also test zero-shot and few-shot prompting;
prompts are in Appendix~\ref{app:prompt_baseline} and results
in Appendix~\ref{app:table4_ci}.

\paragraph{Implementation.}Implementation and reproducibility details
are in Appendix~\ref{reproducibility}.

\section{Results}

\subsection{Discriminative Consistency Bias}
\label{sec:sentiment-results}

Table~\ref{tab:dgi} compares the three metrics, which helps separate dialect effects from other sources of disagreement. Two-way DGI is always lower than
DGI because it also requires the same label for
$\text{AAE}_{\text{back}}$. The gap is small for Gemma-3
($0.849$ vs.\ $0.844$) but large for DeepSeek-R1 and
Llama-3.1-70B (about $0.19$), so these two models often change
their label even between two AAE versions of the same post. cDGI
keeps only pairs where the model gives the same label to AAE and
$\text{AAE}_{\text{back}}$. On these pairs, any remaining
disagreement comes from the SAE version, so it mostly reflects
the change of dialect rather than translation drift.

cDGI is higher than DGI for every model, but only by $0.01$ to
$0.09$, and for four of the six models the increase lies within
the 95\% CI. Translation drift therefore explains only a small
part of the gap: after conditioning, 82\% to 95\% of the AAE--SAE
disagreement remains. For Llama-3.1-70B and DeepSeek-R1, more than
a third of translation-stable pairs still receive different
sentiment labels. In practice, for 13\% to 48\% of
translation-stable posts, the predicted sentiment depends on
whether the post is written in AAE or SAE. cDGI is also stable under alternative conditioning criteria, with the model ordering unchanged
(Appendix~\ref{app:cdgi-sensitivity}).

\begin{table}[t]
\centering
\scriptsize
\setlength{\tabcolsep}{3pt}
\renewcommand{\arraystretch}{1}
\begin{tabular}{lccc}
\toprule
\textbf{Model} & \textbf{DGI} & \textbf{Two-way DGI} & \textbf{cDGI} \\
\midrule
Gemma-3                 & $0.849 \pm 0.005$ & $0.844 \pm 0.005$ & $0.866 \pm 0.032$ \\
Phi-4                   & $0.814 \pm 0.006$ & $0.788 \pm 0.006$ & $0.827 \pm 0.033$ \\
Mistral   & $0.795 \pm 0.006$ & $0.749 \pm 0.006$ & $0.815 \pm 0.030$ \\
Phi-3          & $0.780 \pm 0.006$ & $0.761 \pm 0.006$ & $0.792 \pm 0.033$ \\
DeepSeek-R1             & $0.491 \pm 0.008$ & $0.302 \pm 0.007$ & $0.583 \pm 0.035$ \\
Llama-3.1-70B           & $0.468 \pm 0.008$ & $0.278 \pm 0.007$ & $0.525 \pm 0.031$ \\
\bottomrule
\end{tabular}
\caption{Discriminative consistency on sentiment classification
(mean $\pm$ 95\% CI; higher is better, $1.0$ = perfect dialect
invariance). \textbf{DGI}: AAE--SAE prediction agreement.
\textbf{Two-way DGI}: AAE--SAE--$\text{AAE}_{\text{back}}$
agreement. \textbf{cDGI}: AAE--SAE agreement restricted to
translation-stable examples.}
\label{tab:dgi}
\end{table}

\subsection{Generative Preference Bias}
\label{sec:continuation-results}

We next test whether the preference appears in language modeling
itself, independent of classification, using perplexity and MC
continuation; LP results follow in \S\ref{sec:feature-results}.

\begin{table*}[t]
\centering
\setlength{\tabcolsep}{2.5pt}
\renewcommand{\arraystretch}{1.1}
\newcommand{\ours}[1]{\cellcolor{green!10}#1}
\resizebox{\textwidth}{!}{%
\begin{tabular}{@{}l *{5}{cccc@{\hspace{6pt}}} cccc@{}}
\toprule
 & \multicolumn{4}{c}{\textbf{Phi-4}} & \multicolumn{4}{c}{\textbf{Phi-3}} & \multicolumn{4}{c}{\textbf{Mistral}} & \multicolumn{4}{c}{\textbf{DeepSeek-R1}} & \multicolumn{4}{c}{\textbf{Gemma-3}} & \multicolumn{4}{c}{\textbf{Llama-3.1-70B}} \\
\cmidrule(lr){2-5}\cmidrule(lr){6-9}\cmidrule(lr){10-13}\cmidrule(lr){14-17}\cmidrule(lr){18-21}\cmidrule(lr){22-25}
 & Base & CAD & LoRA & \ours{Ours} & Base & CAD & LoRA & \ours{Ours} & Base & CAD & LoRA & \ours{Ours} & Base & CAD & LoRA & \ours{Ours} & Base & CAD & LoRA & \ours{Ours} & Base & CAD & LoRA & \ours{Ours} \\
\midrule
DGI $\uparrow$ & 0.74 & 0.76 & 0.78 & \ours{\textbf{0.86}} & 0.73 & 0.75 & 0.75 & \ours{\textbf{0.77}} & 0.79 & 0.83 & 0.83 & \ours{\textbf{0.86}} & 0.55 & 0.56 & 0.57 & \ours{\textbf{0.63}} & 0.80 & 0.81 & 0.82 & \ours{\textbf{0.85}} & 0.51 & 0.53 & 0.56 & \ours{\textbf{0.65}} \\
cDGI $\uparrow$ & 0.77 & 0.77 & 0.79 & \ours{\textbf{0.87}} & 0.74 & 0.76 & 0.76 & \ours{\textbf{0.78}} & 0.81 & 0.84 & 0.84 & \ours{\textbf{0.87}} & 0.59 & 0.59 & 0.59 & \ours{\textbf{0.64}} & 0.82 & 0.82 & 0.83 & \ours{\textbf{0.86}} & 0.57 & 0.58 & 0.59 & \ours{\textbf{0.66}} \\
\midrule
AAE PPL $\downarrow$ & 123.9 & 124.8 & 113.0 & \ours{\textbf{107.4}} & 85.8 & 86.1 & 83.7 & \ours{\textbf{82.8}} & 82.8 & 84.8 & 78.1 & \ours{\textbf{77.3}} & 142.8 & 143.4 & 140.0 & \ours{\textbf{133.0}} & 133.4 & 135.6 & 129.3 & \ours{\textbf{121.9}} & 67.7 & 69.1 & 64.9 & \ours{\textbf{64.1}} \\
SAE PPL & 26.7 & 27.8 & 27.4 & \ours{27.9} & 20.6 & 23.5 & 23.1 & \ours{24.6} & 23.3 & 24.8 & 23.7 & \ours{23.9} & 28.1 & 30.5 & 28.8 & \ours{28.3} & 56.3 & 57.7 & 56.1 & \ours{56.6} & 15.5 & 15.8 & 15.6 & \ours{15.7} \\
\midrule
AAE Match $\uparrow$ & 39.3 & 43.8 & 45.1 & \ours{\textbf{58.3}} & 32.6 & 39.4 & 42.7 & \ours{\textbf{56.9}} & 29.5 & 33.5 & 41.6 & \ours{\textbf{50.1}} & 43.8 & 50.2 & 50.9 & \ours{\textbf{66.2}} & 42.9 & 46.2 & 51.8 & \ours{\textbf{61.0}} & 41.3 & 47.2 & 53.7 & \ours{\textbf{65.4}} \\
SAE Match & 89.1 & 89.4 & 88.7 & \ours{88.7} & 79.7 & 79.6 & 78.6 & \ours{78.2} & 88.6 & 88.9 & 88.7 & \ours{88.4} & 75.5 & 73.8 & 73.5 & \ours{72.6} & 82.9 & 83.3 & 83.1 & \ours{83.3} & 88.6 & 88.5 & 87.8 & \ours{87.7} \\
\midrule
LP $\downarrow$ & 60.6 & 61.7 & 55.1 & \ours{\textbf{48.2}} & 70.8 & 72.4 & 64.9 & \ours{\textbf{60.3}} & 57.3 & 58.1 & 53.0 & \ours{\textbf{49.7}} & 68.9 & 70.5 & 66.3 & \ours{\textbf{60.8}} & 89.3 & 92.1 & 86.7 & \ours{\textbf{82.9}} & 34.9 & 37.2 & 31.6 & \ours{\textbf{27.6}} \\
\bottomrule
\end{tabular}}
\caption{Bias mitigation on the held-out test split. \textit{Base}:
unsteered model. \textit{CAD}: context-aware decoding
\citep{shi-etal-2024-trusting}. \textit{LoRA}: LoRA fine-tuning
\citep{hu2022lora} on dialect-matched continuations from the
direction split. \textit{Ours}: activation steering (green).
AAE/SAE Match: \% of MC items where the model picks the
dialect-matched continuation under AAE/SAE context.
$\uparrow$/$\downarrow$: higher/lower is better; \textbf{bold}: best
value per model. SAE PPL and SAE Match measure side effects on SAE,
should stay close to \textit{Base}, and are not bolded. Prompting
baselines and 95\% CIs for all methods are in
Table~\ref{tab:mitigation_ci} (Appendix~\ref{app:table4_ci}); full
$\beta$ sweep in Appendix~\ref{app:full-steering}.}
\label{tab:main-debiasing-results}
\end{table*}

\begin{figure}[t]

    \centering
    \includegraphics[width=0.95\linewidth]{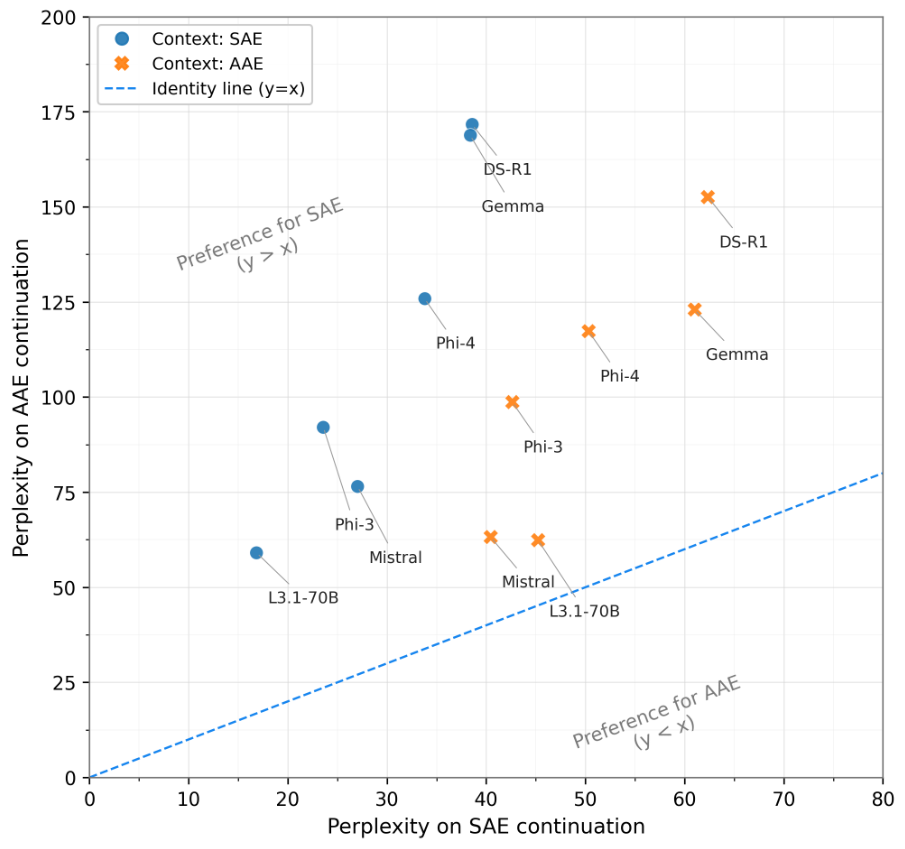}
  
\caption{Continuation perplexity for SAE ($x$-axis) and AAE
($y$-axis) continuations, by model and context. Points above
$y = x$ indicate SAE preference.}
    \label{fig:perlexity-scatter}
\end{figure}

\begin{figure}[t]
    \centering
    \includegraphics[width=0.95\linewidth]{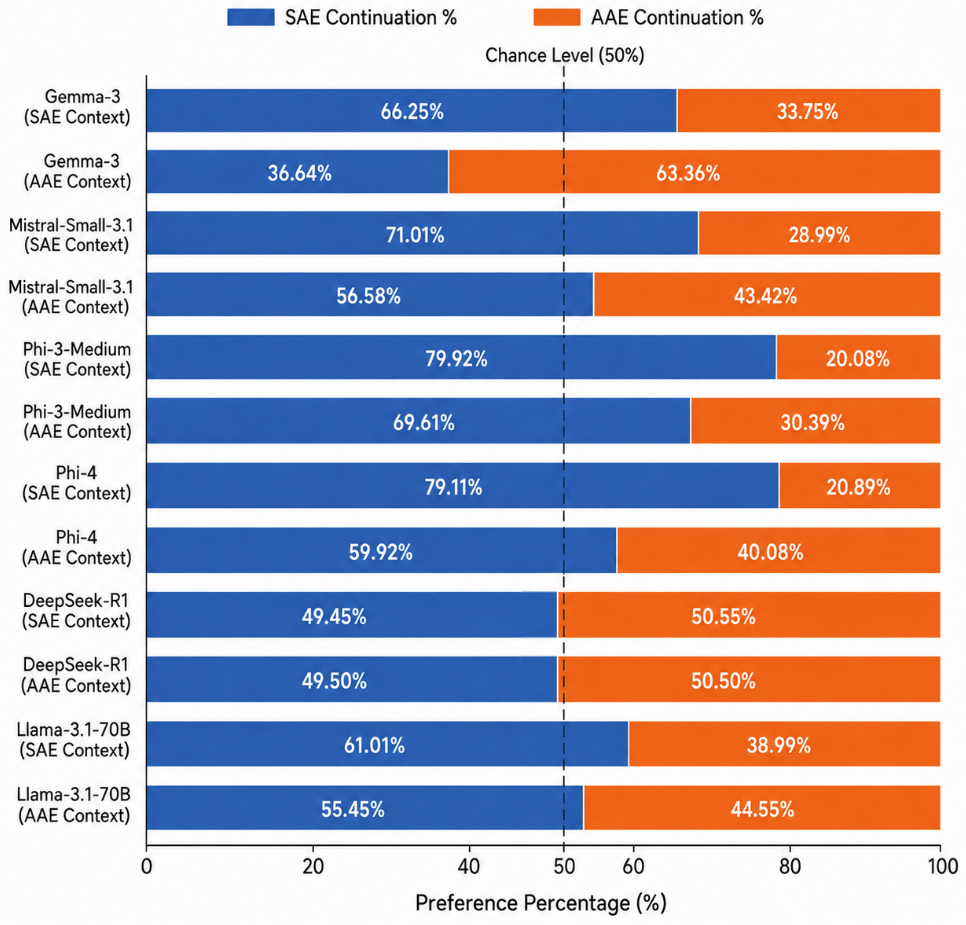}
\caption{Multiple-choice continuation preferences under SAE and
AAE contexts. Bars show the \% selecting each continuation;
dashed line = 50\% chance. }
    \label{fig:mc-stacked-bar}
\end{figure}

\paragraph{Perplexity Analysis.}
Fig.~\ref{fig:perlexity-scatter} reports continuation perplexity
for all four context$\rightarrow$continuation combinations. Under
SAE context, all six models assign lower perplexity to SAE
continuations than to AAE: the expected pattern when continuation
matches context. \uline{Under AAE context, however, every model
still assigns lower perplexity to SAE continuations.} Models
adapt to SAE context but fail to adapt to AAE context,
defaulting to SAE regardless of input dialect.

\paragraph{Multiple-Choice Evaluation.}
Fig.~\ref{fig:mc-stacked-bar} confirms this asymmetry in the
MC task: most models still prefer SAE even when the context
is AAE: Phi-4 selects SAE $59.9\%$ of the time
under AAE context (vs.\ $40.1\%$ for AAE). Two models deviate.
Gemma-3 reverses the pattern, preferring AAE under AAE context
($63.4\%$), while DeepSeek-R1 stays near $50\%$ in both contexts,
suggesting weak context-sensitivity rather than active
dialect-matching. Table~\ref{tab:pmc-example} in
Appendix~\ref{app:qualitative-steering} shows typical failures.



\subsection{Feature-Level Bias Localization}
\label{sec:feature-results}

\begin{figure}[t]
    \centering
    \includegraphics[width=\linewidth]{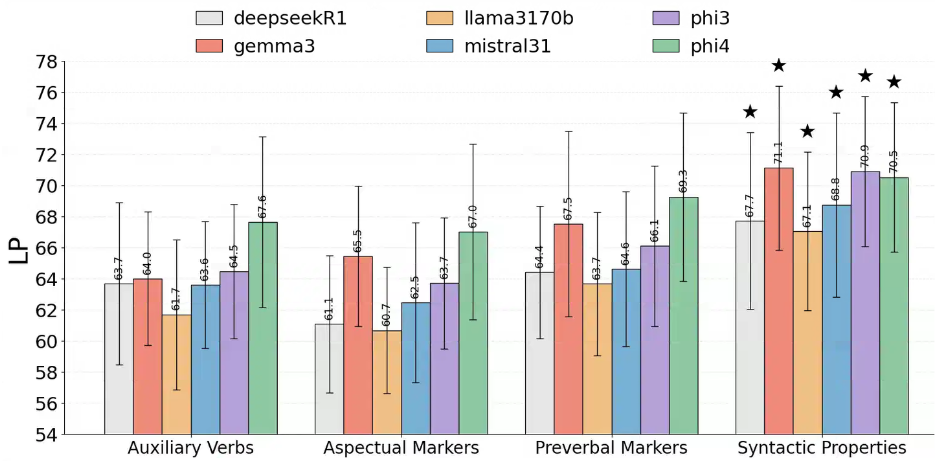}
    \caption{Feature-level bias localization. For each model,
    bars show the average $\mathrm{LP}$ bias (95\% CIs)
    across four AAE feature classes; higher $\mathrm{LP}$
    indicates stronger SAE preference. $\star$ marks the highest
    $\mathrm{LP}$ per model.}
    \label{fig:feature_level_bias}
\end{figure}

Fig.~\ref{fig:feature_level_bias} shows which AAE features most
strongly trigger the bias, reporting the average $\mathrm{LP}$ for
each of the four feature classes; higher $\mathrm{LP}$ indicates
stronger SAE preference.

All four feature classes yield positive $\mathrm{LP}$ across all
six models: SAE preference is not tied to any single marker but
spans a broad range of AAE forms. Syntactic properties, especially negative concord
(e.g., ``ain't nobody'', ``can't nobody''), yield the highest
$\mathrm{LP}$ for every model tested. Preverbal
markers also produce relatively strong effects, while auxiliary
and aspectual markers show the same direction but smaller
magnitudes.

\subsection{Bias Mitigation}
\label{sec:steering-results}

We evaluate activation steering as a test-time mitigation method
against \textit{Base}, \textit{CAD}, and \textit{LoRA}
(\S\ref{sec:setup}); LoRA is the only method that requires
training. Table~\ref{tab:main-debiasing-results} reports DGI, cDGI, PPL, MC, and LP across all six models, with steering strength $\beta$ selected per model on validation set to balance bias reduction and language modeling ability; the full sweep and bias--utility frontier appear in Appendices~\ref{app:full-steering} and~\ref{app:tradeoff}. Steering uses separate splits (\S\ref{sec:setup}), so absolute values are not directly comparable to \S\S\ref{sec:sentiment-results}--\ref{sec:feature-results}.

\uline{Activation steering reaches the best value on every bias metric for all six models}, while largely preserving SAE fluency. 

\paragraph{DGI and cDGI.}
Both metrics improve under steering for every model (e.g.,
Phi-4: DGI $0.74 \rightarrow 0.86$, cDGI $0.77 \rightarrow
0.87$). Steering raises DGI by $0.04$ to $0.14$, while CAD and LoRA
gain at most $0.05$. Still, DGI stays at or below $0.86$, so
dialect-dependent labeling is reduced but not removed. cDGI here uses the 578 annotator-judged-natural test sentences.

\paragraph{Perplexity (PPL).}
Because perplexity scales differ across models, we report
relative changes from \textit{Base}. Steering
lowers AAE perplexity for every model, by 3.5\% (Phi-3) to
13.3\% (Phi-4), more than LoRA (2.0\% to 8.8\%), while CAD slightly
raises it. SAE perplexity rises by less than 5\% for five models
(Phi-4 $+4.5\%$, Mistral $+2.6\%$, DeepSeek-R1 $+0.7\%$, Gemma-3
$+0.5\%$, Llama-3.1-70B $+1.3\%$) and by 19.4\% for Phi-3.

\paragraph{General capability.}
We find no evidence that steering degrades capabilities
unrelated to dialect. At the selected $\beta$, accuracy on
MMLU \citep{hendrycks2021measuring} and
GSM8K \citep{cobbe2021gsm8k} changes by at most 2.5 percentage points in either direction,
within the sampling error of the evaluation subsets
(Appendix~\ref{app:capability}).

\paragraph{Multiple-choice preference (MC).}
Steering raises AAE Match by 18 to 24 points
(e.g., Phi-4: $39.3\% \rightarrow 58.3\%$), while SAE Match
stays within 3 points of \textit{Base}. LoRA is the
strongest baseline (6 to 12 points), and steering exceeds it by 8
to 15 points on every model. After steering, AAE
Match ranges from 50.1\% to 66.2\%, so models still often choose
the SAE continuation under AAE context.

\paragraph{Log-probability bias (LP).}
LP decreases for every model under steering, by 7\% to 21\%
(e.g., Phi-4: $60.6 \rightarrow 48.2$). LoRA lowers LP by 3\% to
10\%, and steering reaches a lower LP than LoRA on all six models.
CAD slightly increases LP, by 1\% to 7\%.

\paragraph{Human evaluation of generations.}
Three native AAE speakers blindly compared base and steered
continuations for 200 AAE contexts. The steered one was judged a
better dialect match (73.0\% vs.\ 9.8\%) and more natural (63.3\%
vs.\ 10.6\%), and was less often flagged as exaggerated (5.1\% vs.\
35.2\%; Appendix~\ref{app:generation}).

Zero- and few-shot prompting are weaker than steering on every
bias metric (Table~\ref{tab:mitigation_ci}), and ablations on prompt
wording, layers, and $\beta$ give consistent results
(Appendices~\ref{app:prompt-ablation}--\ref{app:full-steering}).

\section{Conclusion}


We showed that all six open-weight LLMs we tested systematically
favor Standard American English over African American English
(AAE), even when the user is writing in AAE. Our contributions
are: \textsc{Real-AAE}, the largest real-AAE parallel corpus to
date; cDGI and a feature-level analysis identifying syntactic
properties as the strongest trigger of dialect preference; and
the first application of activation steering to dialect bias in AAE, which outperforms prompting, CAD, and LoRA fine-tuning without training. The framework requires only a feature lexicon and
parallel AAE--SAE pairs to extend to other marginalized
varieties.

\section*{Limitations}

\paragraph{Source data and register.} \textsc{Real-AAE} draws
from the TwitterAAE corpus
\citep{blodgett-etal-2016-demographic}, which identifies AAE-speaker
tweets via demographic inference. This provides naturally occurring
AAE at scale but captures the written social-media register, which
may not fully generalize to spoken, formal, or other written
registers. Extending the framework to additional registers is
straightforward and a natural direction for future work.

\paragraph{LLM-generated SAE translations.} Our SAE translations
are produced by Gemini-3-flash and validated both automatically
(BERTScore F1 = 0.95) and by three native AAE speakers. Our cDGI
metric is specifically designed to isolate model bias from
most residual translation drift but not remove it in the conditioning examples.

\paragraph{Annotator regional coverage.} Our human validation
involves three native AAE speakers from the West Coast and
Northeast/Mid-Atlantic. Because annotators validate corpus quality
(translation faithfulness and AAE naturalness) rather than the
bias findings themselves, their regional coverage does not affect
the LLM-behavior results we report; it does, however, shape the
naturalness judgments used to construct the cDGI conditioning set.
Expanding the annotator pool to include Southern AAE speakers is
an important next step.

\section*{Ethical Considerations}

This work aims to reduce the disparities and dialect erasure that AAE speakers face in language technologies. We do not treat AAE as erroneous or use dialect as a proxy for racial identity.
The tweets come from the TwitterAAE corpus \citep{blodgett-etal-2016-demographic}, which is distributed for non-commercial academic use. We release the triplets with user mentions and URLs removed. The corpus and feature lexicon contain profanity and slurs as they occur in the source data. We include them without endorsement.

Three native AAE speakers on the research team independently validated 2,000 examples in randomized order, without seeing model outputs or automatic validation signals, and later compared base and steered continuations without knowing which was steered (Appendix~\ref{app:generation}). Both annotation tasks are documented in Appendix~\ref{app:annotation-guidelines}, including the content warning shown to annotators. The annotators are included as co-investigators. Our institution determined that this annotation work did not constitute human-subjects research requiring IRB review.
Steering is intended for auditing and mitigation. Reversing the direction could amplify bias. AAE is not monolithic, and our data and lexicon do not represent its full diversity. We encourage deployment in consultation with affected communities, with attention to user consent and dialect accommodation.

\bibliography{references}

\appendix
\section{Comparison with Prior Work}
\label{app:prior-work}

Table~\ref{tab:comparison} compares \textsc{Real-AAE} with prior
resources for evaluating and mitigating dialect bias
(\S\ref{sec:related-work}). We group prior work by how the AAE
text is obtained. \textit{SAE-first} resources start from SAE and
derive AAE from it, either with rules, with an LLM, or by asking
speakers to rewrite SAE prompts, so the resulting AAE is not
naturally occurring. \textit{AAE-first} resources start from AAE
written by AAE speakers and pair it with an SAE counterpart.
\textsc{Real-AAE} is the only resource that combines naturally
occurring AAE, native-speaker validation, both discriminative and
generative auditing, feature-level analysis, and a mitigation
method.

\newcommand{\cmark}{\ding{51}}
\newcommand{\xmark}{\textcolor{gray}{\ding{55}}}

\begin{table*}[t]
\centering
\footnotesize
\setlength{\tabcolsep}{4pt}
\renewcommand{\arraystretch}{1.2}
\begin{tabularx}{\linewidth}{@{}l Y l r c c c c@{}}
\toprule
& \multicolumn{3}{c}{\textbf{Data}}
& \multicolumn{3}{c}{\textbf{Bias auditing}}
& \\
\cmidrule(lr){2-4} \cmidrule(lr){5-7}
\textbf{Work}
& \textbf{How AAE/SAE pairs are built}
& \textbf{Validation}
& \textbf{Size}
& \textbf{Discrim.}
& \textbf{Generative}
& \makecell{\textbf{Feature-}\\\textbf{level}}
& \textbf{Mitigation} \\
\midrule
\multicolumn{8}{@{}l}{\textit{SAE-first: AAE is derived from SAE text}} \\
\makecell[l]{VALUE\\\citep{ziems2022value}}
  & Rule-based SAE$\rightarrow$AAE & Human & 2,880$^\dagger$
  & Task acc. & \xmark & \cmark & \xmark \\[3pt]
\makecell[l]{Multi-VALUE\\\citep{ziems-etal-2023-multi}}
  & Rule-based SAE$\rightarrow$dialects & Native & 7,983$^\dagger$
  & Task acc. & \xmark & \cmark & Data aug. \\[3pt]
\makecell[l]{AAVENUE\\\citep{gupta2024aavenue}}
  & LLM SAE$\rightarrow$AAE & Human & 5,000
  & Task acc. & \xmark & \xmark & \xmark \\[3pt]
\makecell[l]{EnDive\\\citep{gupta2025endive}}
  & LLM SAE$\rightarrow$dialects & Native & 40k+
  & Reas.\ acc. & \xmark & \xmark & \xmark \\[3pt]
\makecell[l]{ReDial\\\citep{lin-etal-2025-assessing}}
  & Speakers rewrite SAE prompts & Native & 1,200+
  & Reas.\ acc. & \xmark & \xmark & \xmark \\[3pt]
\midrule
\multicolumn{8}{@{}l}{\textit{AAE-first: naturally occurring AAE}} \\
\citet{groenwold2020}
  & Crowd-written SAE counterparts & -- & 2,019
  & \xmark & \cmark & \xmark & \xmark \\[3pt]
\citet{deas-etal-2023-evaluation}
  & Speaker-written WME counterparts & Native & 346
  & \xmark & \cmark & \xmark & \xmark \\[3pt]
\citet{hofmann2024dialect}
  & Matched guise & -- & 2,019
  & Trait assoc. & \xmark & \cmark & \xmark \\[3pt]
\citet{fleisig2024linguisticbiaschatgptlanguage}
  & Prompt-elicited responses & -- & $\sim$3,000
  & \xmark & \cmark & \cmark & \xmark \\[3pt]
\citet{mire2025rejecteddialectsbiasesafrican}
  & Preference pairs & -- & \makecell[r]{1,843\\+ 2,365}
  & RM score & \xmark & \xmark & \xmark \\[3pt]
\citet{Hassan_Khattak_Seyyed-Kalantari_2025}
  & Back-translation & -- & $\sim$5,000
  & DGI & \xmark & \xmark & \xmark \\[3pt]
\midrule
\rowcolor{green!10}
\textbf{\textsc{Real-AAE} (Ours)}
  & \textbf{LLM translation + back-translation} & \makecell[l]{\textbf{Native +}\\\textbf{automatic}} & \textbf{17,479}
  & \textbf{DGI, cDGI} & \textbf{PPL, LP, MC} & \cmark & \textbf{Steering} \\
\bottomrule
\end{tabularx}
\caption{Comparison with prior work on dialect bias evaluation and
mitigation, grouped by how the AAE text is obtained.
\textbf{Validation}: Human = validated by human annotators;
Native = validated or written by native AAE speakers; automatic
= automatic content- and sentiment-preservation checks; -- = not reported. \textbf{Size}: number of
AAE examples; $^\dagger$ evaluation subset rather than full
benchmark size. \textbf{Discrim.} (discriminative): metric for consistency
of classification or scoring across dialects (acc.: accuracy; Reas.: reasoning;
Trait assoc.: trait association; RM: reward model; DGI/cDGI:
(conditional) Dialect Group Invariance). \textbf{Generative}:
whether bias in generation or continuation is measured (PPL:
perplexity; LP: log-probability; MC: multiple-choice
continuation). \textbf{Feature-level}: whether bias is
attributed to specific AAE features. WME: White Mainstream
English; Data aug.: data augmentation.}
\label{tab:comparison}
\end{table*}

\section{Additional Quality Validation Results}
\label{app:quality_validation}

\subsection{Automatic Validation}
\label{app:auto-validation}
\begin{figure}[h]
    \centering
    \includegraphics[width=0.75\linewidth]{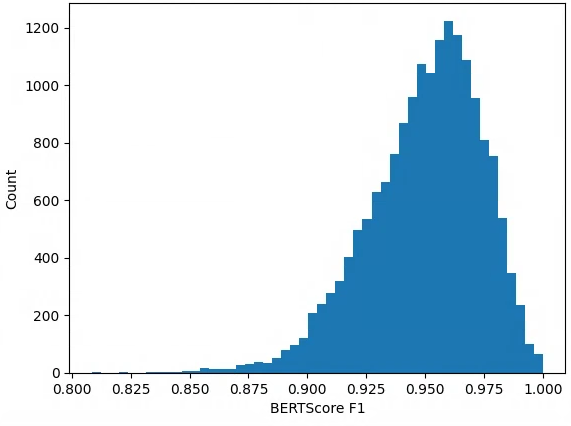}
    \caption{Distribution of BERTScore F1 between original AAE sentences and their back-translated AAE versions, computed with cardiffnlp/twitter-roberta-base-2022-154m. Scores are concentrated near 0.95 and above, suggesting strong semantic preservation in the back-translation step.}
    \label{fig:bertscore-hist}
\end{figure}

This appendix provides additional results for the automatic validation described in \S\ref{construction}. Table~\ref{tab:sentiment-confusion} shows the sentiment confusion matrix between AAE and SAE sentences, and Fig.~\ref{fig:bertscore-hist} shows the distribution of BERTScore F1 between original AAE sentences and their back-translated versions.

\begin{table}[h]
\centering
\small
\setlength{\tabcolsep}{6pt}
\renewcommand{\arraystretch}{1.1}
\begin{tabular}{lccc}
\toprule
\textbf{AAE} $\backslash$ \textbf{SAE} & \textbf{Negative} & \textbf{Neutral} & \textbf{Positive} \\
\midrule
Negative & 7268 & 848 & 357 \\
Neutral  & 833  & 3832 & 949 \\
Positive & 87   & 171  & 3134 \\
\bottomrule
\end{tabular}
\caption{Sentiment consistency confusion matrix between AAE and SAE sentences, measured with cardiffnlp/twitter-roberta-base-sentiment-latest. Rows denote sentiment labels predicted for AAE, and columns denote sentiment labels predicted for SAE. Most mass lies on the diagonal, indicating that sentiment is usually preserved across dialectal variants.}
\label{tab:sentiment-confusion}
\end{table}

\subsection{Translation Filtering and Error Analysis}
\label{app:translation-errors}

Translation failures were rare. Of 17,495 AAE-to-SAE translation
attempts, 16 were excluded
(Table~\ref{tab:translation-errors}). Ten were blocked by the
translator's safety filter, chiefly for sexual content
referencing minors or for violent insults, and we do not
reproduce these examples. Five were removed because the
translator appended explanatory commentary instead of a clean
translation, and one returned empty output.

\begin{table}[h]
\centering
\small
\setlength{\tabcolsep}{6pt}
\renewcommand{\arraystretch}{1.1}
\begin{tabular}{lr}
\toprule
\textbf{Outcome} & \textbf{Count} \\
\midrule
Retained                                  & 17,479 \\
Excluded: translator safety block         & 10 \\
Excluded: translator-added explanation    & 5 \\
Excluded: empty translator output         & 1 \\
\midrule
Total translation attempts                & 17,495 \\
\bottomrule
\end{tabular}
\caption{Filtering outcomes in the translation stage of the
\textsc{Real-AAE} construction pipeline.}
\label{tab:translation-errors}
\end{table}

For example, one removed output rendered the translation
correctly but appended a parenthetical note explaining the
origin of a slang term: \emph{[I don’t even know why he is trying to call me this late, lol] (Note: ‘ion’ is an informal spelling of ‘I don’t)}.

\section{Additional Methodological Details}
\label{sec:method-appendix}

\subsection{SAE-Guided Semantic Segmentation}
\label{app:splitting}

\begin{figure*}[t]
    \centering
    \includegraphics[width=0.75\linewidth]{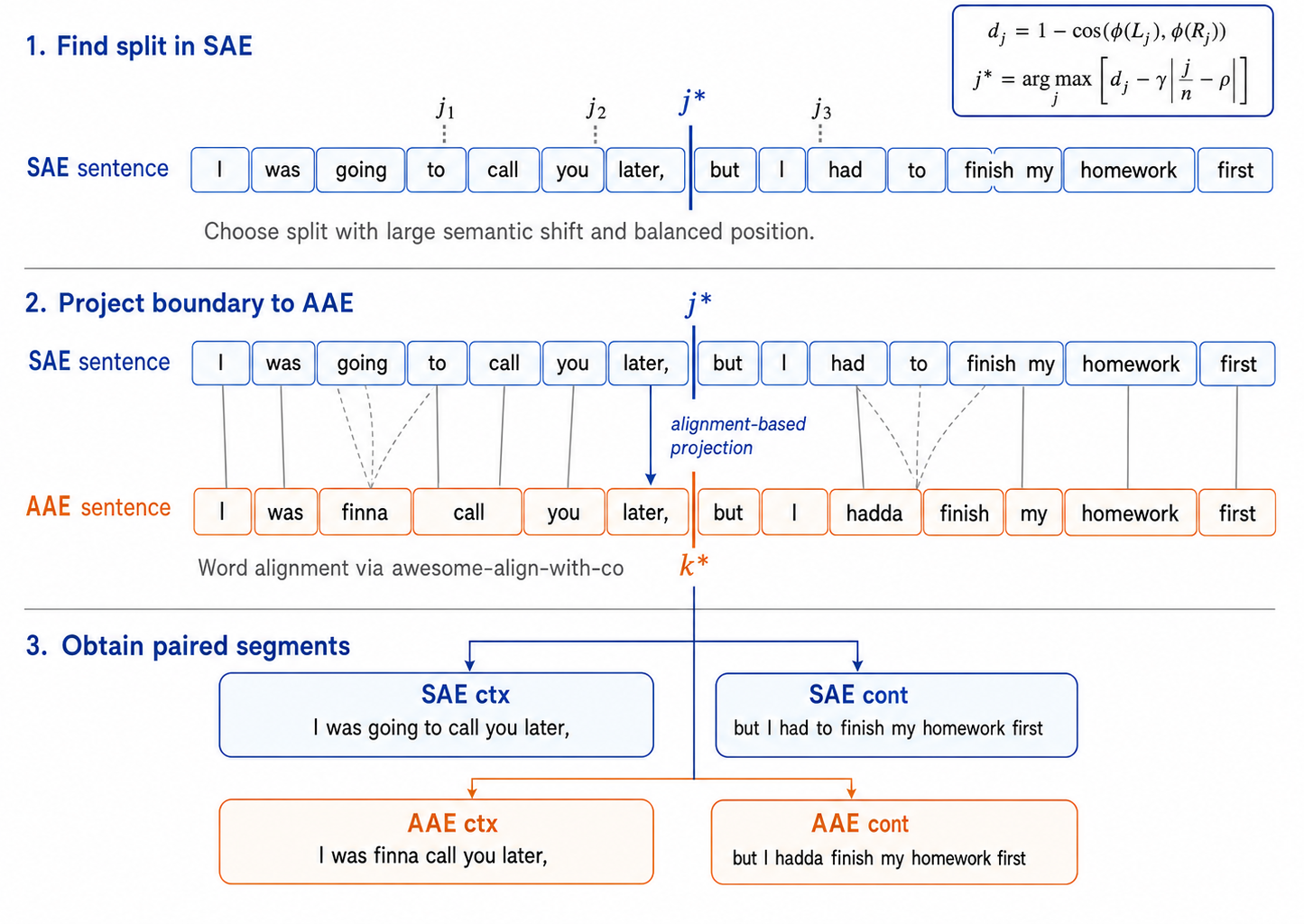}
    \caption{Alignment-based SAE-guided semantic segmentation.
First, we select a context--continuation boundary $j^*$ in the SAE sentence using semantic shift and a balance penalty.
Second, we project this boundary to the paired AAE sentence using word-level alignment from aneuraz/awesome-align-with-co, obtaining the aligned AAE boundary $k^*$.
The final output is a set of paired SAE and AAE context-continuation segments.}
    \label{fig:sentence-split}
\end{figure*}

Fig.~\ref{fig:sentence-split}  illustrates our alignment-based segmentation procedure. 
The goal is to split each paired SAE--AAE sentence into matched context and continuation segments.

We first identify the split point in the SAE sentence. Given the SAE sentence 
``I was going to call you later, but I had to finish my homework first,'' 
we consider several candidate boundaries near the middle of the sentence, such as $j_1$, $j_2$, and $j_3$. 
For each candidate $j$, we compute the semantic shift between the left segment $L_j$ and the right segment $R_j$:
\[
d_j = 1-\cos(\phi(L_j),\phi(R_j)).
\]
We then select the boundary with a large semantic shift while penalizing highly imbalanced splits:
\begin{equation}
j^* = \arg\max_j \left[d_j - \gamma \left| \frac{j}{n} - \rho \right| \right]
\label{eq:split_obj}
\end{equation}
In Fig.~\ref{fig:sentence-split}, this selects the boundary after ``later,'' in the SAE sentence.

In Eq.~\ref{eq:split_obj} , we set $\rho$ = 0.5 to encourage roughly balanced context--continuation splits. We use $\gamma$ = 0.3 as a fixed balance penalty to trade off semantic shift against split length imbalance. This avoids boundaries that are too close to either end of the sentence while still favoring semantically meaningful splits.
We then project this SAE boundary to the paired AAE sentence using word-level alignment. 
Instead of mapping the boundary by the same relative position, we align SAE and AAE words using aneuraz/awesome-align-with-co. 
The aligned words around $j^*$ are used to choose the corresponding AAE boundary $k^*$. 
In the example, the SAE context 
``I was going to call you later,'' 
is aligned with the AAE context 
``I was finna call you later,'' 
while the SAE continuation 
``but I had to finish my homework first'' 
is aligned with the AAE continuation 
``but I hadda finish my homework first.''

This procedure avoids assuming that SAE and AAE have the same length or word order. 
It instead projects the split through word-level correspondence, which gives cleaner paired context--continuation segments for the continuation preference experiments.

\subsection{Segmentation Robustness}
\label{app:seg-robustness}

Our continuation experiments depend on the alignment-based
segmentation of Appendix~\ref{app:splitting}, so we test whether
the audit findings change under different split choices. We
compare three methods on the same pairs: our semantic-shift
split, a fixed midpoint split at 50\% of SAE tokens, and a random
split drawn uniformly between 40\% and 60\% of SAE tokens. The midpoint and random boundaries are projected to the AAE sentence with the same word alignment as our method.

Table~\ref{tab:seg-lp} shows that LP is stable. It stays positive
for every model under every split, the model ordering is
unchanged, and most values move by less than 3 points. The core
audit finding, that every model prefers SAE continuations even
under AAE context, does not depend on how we segment.

MC behaves differently (Table~\ref{tab:seg-mc}). For most models,
match rates rise sharply under the alternative splits, on both the
SAE and the AAE side (e.g., Phi-4 AAE Match $40.1 \rightarrow 60.1$, DeepSeek-R1
SAE Match $49.5 \rightarrow 67.9$). This is a seam artifact, not
a change in bias. The matched continuation is the second half of
the same sentence, so it joins the context cleanly wherever we
cut. The mismatched continuation comes from the parallel
sentence, and when the two split points do not line up, it picks
up missing or repeated words at the seam. Models then avoid it
for grammatical reasons unrelated to dialect, which inflates both
match rates and understates the bias. LP is largely immune
because the seam penalty appears in both directions and cancels
in the difference. We therefore treat LP as the primary
continuation metric and read MC as a confirmatory signal under a
fixed segmentation. Comparisons between mitigation methods are
unaffected, since all methods are scored on identical items under
the same segmentation.

\begin{table}[h]
\centering
\footnotesize
\setlength{\tabcolsep}{4pt}
\renewcommand{\arraystretch}{1.05}
\resizebox{\columnwidth}{!}{
\begin{tabular}{lcccccc}
\toprule
Split & Gemma-3 & Phi-4 & Mistral & Phi-3 & \makecell{DeepSeek\\-R1} & \makecell{Llama-3.1\\-70B} \\
\midrule
Ours     & 89.29 & 60.64 & 57.26 & 70.79 & 68.89 & 34.86 \\
Midpoint & 91.62 & 62.36 & 58.12 & 70.93 & 70.31 & 39.07 \\
Random   & 90.85 & 62.91 & 57.99 & 71.42 & 69.54 & 37.62 \\
\bottomrule
\end{tabular}
}
\caption{LP under alternative segmentations. LP stays positive
for every model under every split and the model ordering is
unchanged.}
\label{tab:seg-lp}
\end{table}

\begin{table}[h]
\centering
\footnotesize
\setlength{\tabcolsep}{4pt}
\renewcommand{\arraystretch}{1.05}
\resizebox{\columnwidth}{!}{
\begin{tabular}{lcccccc}
\toprule
Split & Gemma-3 & Phi-4 & Mistral & Phi-3 & \makecell{DeepSeek\\-R1} & \makecell{Llama\\-70B} \\
\midrule
Ours     & 66.3\,/\,63.4 & 79.1\,/\,40.1 & 71.0\,/\,43.4 & 79.9\,/\,30.4 & 49.5\,/\,50.5 & 61.0\,/\,44.6 \\
Midpoint & 79.6\,/\,64.6 & 83.0\,/\,60.1 & 74.7\,/\,58.7 & 83.3\,/\,55.3 & 67.9\,/\,66.4 & 70.5\,/\,64.8 \\
Random   & 77.3\,/\,63.1 & 80.8\,/\,54.6 & 73.9\,/\,55.2 & 81.5\,/\,50.8 & 62.0\,/\,57.7 & 64.1\,/\,54.2 \\
\bottomrule
\end{tabular}
}
\caption{MC match rates (SAE Match\,/\,AAE Match, \%) under
alternative segmentations. }
\label{tab:seg-mc}
\end{table}

\subsection{Prompt Template}
\label{app:prompt}

The prompt template is shown below. We use a minimal structure that avoids instructions about dialect, tone, or style, isolating the model's intrinsic continuation preference. Option ordering (SAE vs.\ AAE as option A or B) is randomized to control for positional bias.

\begin{promptbox}[Multiple-choice evaluation prompt]
\small\ttfamily
Pick the continuation that best follows the context.
Answer only "A" or "B".\\[5pt]
Context: \{context\_text\}\\[3pt]
A: \{A\_continuation\}\\
B: \{B\_continuation\}
\end{promptbox}

\subsection{AAE Grammatical Feature Lexicon}
\label{app:lexicon}

Table~\ref{tab:aae_features} presents the AAE grammatical markers used in our feature-level analysis, adapted from \citet{harris2022exploring}. Features are categorized into four classes: Auxiliary (non-standard auxiliary constructions), Aspectual (habitual and completive markers), Preverbal (pre-verbal lexical items), and Syntactic (clause-level constructions such as negative concord).

\begin{table}[h]
\centering
\small
\begin{tabular}{llll}
\toprule
\textbf{Auxiliary} & \textbf{Aspectual} & \textbf{Preverbal} & \textbf{Syntactic} \\
\midrule
we was        & I be        & aint        & cant nobody \\
they was      & he be       & ain't       & can't nobody \\
finna         & they be     & steady      & he don't \\
tryna         & she be      & stay        & she don't \\
imma          & I been      & he done     & don't never \\
i'mma         & he been     & she done    & he dont \\
bitches was   & she been    & they done   & she dont \\
niggas was    & they been   & yall done   & dont never \\
yall was      & it be       & y'all done  & yall don't \\
y'all was     & niggas be   & you done    & y'all don't \\
you was       & bitches be  & u done      & aint nothing \\
wanna         & yall be     &             & ain't nothing \\
gonna         & y'all be    &             & aint nobody \\
ima           & you be      &             & ain't nobody \\
ion           & u be        &             & yall dont \\
u was         &             &             &  \\
iont          &             &             &  \\
\bottomrule
\end{tabular}
\caption{AAE grammatical feature lexicon adapted from \citet{harris2022exploring}.}
\label{tab:aae_features}
\end{table}

\subsection{Reproducibility Details}
\label{reproducibility}

\begin{figure*}[!t]
    \centering
    \includegraphics[width=\linewidth]{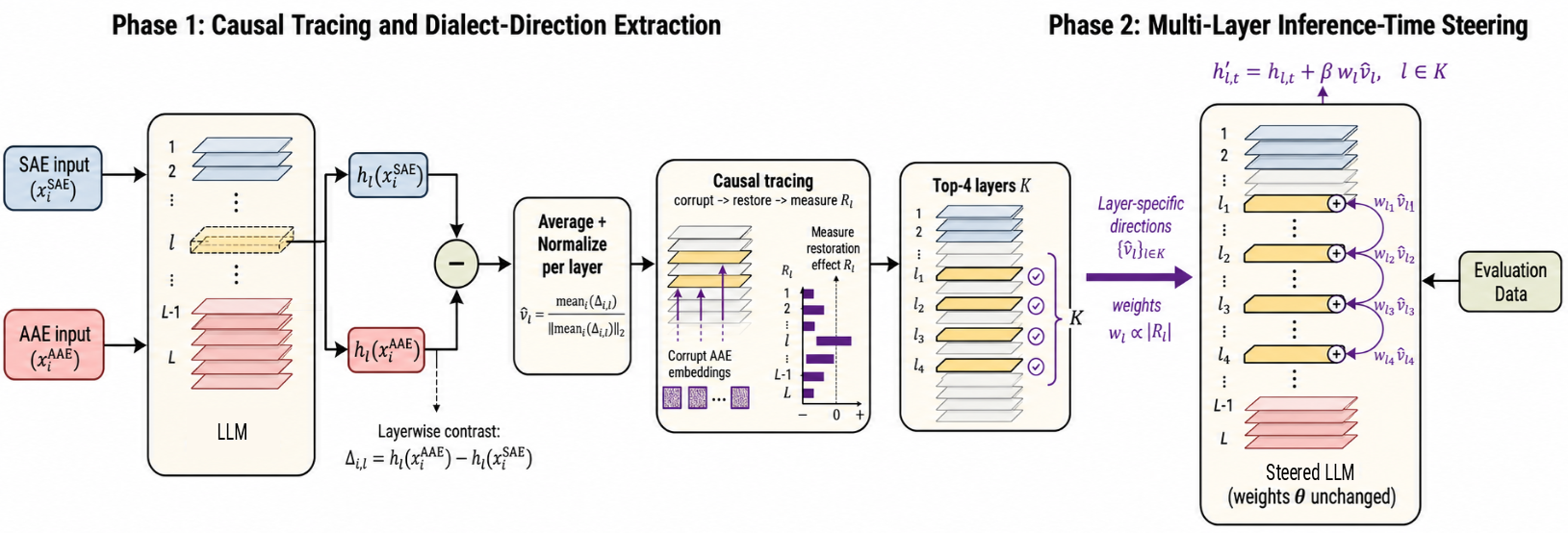}
    \caption{Overview of dialect debiasing with activation steering.
    \textbf{Phase~1:} from paired AAE--SAE inputs, we compute layerwise
    hidden-state contrasts, derive a per-layer dialect direction, and select
    the top-4 causally important layers via corruption-and-restore analysis.
    \textbf{Phase~2:} at inference, we inject these directions into the
    selected layers to steer generation toward dialect-matched behavior,
    without updating model weights.}
    \label{fig:bias-mitigation}
\end{figure*}

All experiments are in PyTorch. We use
vLLM \cite{kwon2023efficient} for inference and Hugging Face Transformers
\cite{wolf2020transformers} for steering, which requires direct access to
residual activations. Perplexity is computed via autoregressive
likelihood; multiple-choice evaluation extracts log-probabilities
and selects the higher-probability continuation. Computations were
performed on NVIDIA H100 80GB GPUs.

All causal-tracing and steering experiments were implemented in PyTorch with Hugging Face \texttt{transformers}. For each model, we used 1,200 AAE-feature sentences for layer selection, 7,000 AAE--SAE pairs for dialect-direction extraction, a
1,000-pair validation set for selecting the steering strength,
and a 2,000-pair test set for all reported results. Causal tracing corrupted the AAE feature-token embeddings with Gaussian noise $\epsilon \sim \mathcal{N}(0,\sigma^2 I)$, where $\sigma = 3.0 \times \mathrm{std}(E_{\mathrm{feat}})$ for the selected feature embeddings in that example, and restored clean activations layer-by-layer at the same token positions. For steering, we extracted one unit dialect direction per selected layer from the last context-token hidden-state contrast, ranked layers by mean restoration effect, and used the top $4$ layers in the multi-layer setting with weights proportional to the absolute restoration effect. We swept $\beta \in \{0, 0.1, 0.2, 0.4, 0.6, 0.8, 1.0\}$ on the
validation set, froze the selected value, and then scored the
test set once.

\section{Supplementary Steering Analyses}
\label{app:steering-analyses}

\subsection{Prompting Baselines}
\label{app:prompt_baseline}

For the zero-shot baseline, we use the original multiple-choice
continuation prompt and add the instruction shown below.

\begin{promptbox}[Zero-shot instruction]
\small\ttfamily
Treat African American English (AAE) and Standard American
English (SAE) as semantically equivalent dialects, and base your
judgment only on the underlying meaning rather than
dialect-specific wording.
\end{promptbox}

For the few-shot baseline, we prepend four exemplars to the same
prompt. Each exemplar shows a context and two candidate
continuations, with the dialect-matched continuation marked as
the correct answer. Two exemplars use AAE contexts and two use
SAE contexts, and the correct letter is balanced between A and B.
All exemplars are drawn from the direction-extraction split and
do not overlap with the validation or test sets.

\begin{promptbox}[Few-shot prompt]
\small\ttfamily
Pick the continuation that best follows the context.
Answer only "A" or "B".\\[5pt]
Context: I was finna call you later,\\
A: but I hadda finish my homework first.\\
B: but I had to finish my homework first.\\
Answer: A\\[5pt]
Context: I was going to call you later,\\
A: but I hadda finish my homework first.\\
B: but I had to finish my homework first.\\
Answer: B\\[5pt]
Context: She said she done told\\
A: him twice already and he still ain't listen.\\
B: him twice already and he still did not listen.\\
Answer: A\\[5pt]
Context: She said she had already told\\
A: him twice already and he still ain't listen.\\
B: him twice already and he still did not listen.\\
Answer: B\\[5pt]
Context: \{context\_text\}\\
A: \{A\_continuation\}\\
B: \{B\_continuation\}
\end{promptbox}

\subsection{Prompt Wording Ablation}
\label{app:prompt-ablation}

The word \emph{best} in our prompt could favor SAE by implying correctness. We therefore replace it with \emph{most likely continuation} and \emph{most stylistically consistent continuation} and rerun the evaluation for all models.

Table~\ref{tab:prompt_ablation} shows that the pattern holds under all three wordings. For Phi-4, Phi-3, Mistral, and Llama-3.1-70B, AAE continuation rates under SAE context stay at 20--40\%, whereas SAE continuation rates under AAE context stay at 50--70\%. The stylistic wording lowers the SAE rate under AAE context for Phi-3, Mistral, and Llama-3.1-70B, but not for Phi-4. Gemma-3 and DeepSeek-R1 keep the patterns in Fig.~\ref{fig:mc-stacked-bar}. The bias therefore does not come from the word \emph{best} alone.

\begin{table}[t]
\centering
\scriptsize
\setlength{\tabcolsep}{4pt}
\renewcommand{\arraystretch}{1.05}
\resizebox{\columnwidth}{!}{
\begin{tabular}{llcccc}
\toprule
& & \multicolumn{2}{c}{SAE ctx} & \multicolumn{2}{c}{AAE ctx} \\
\cmidrule(lr){3-4}\cmidrule(lr){5-6}
Model & Strategy & SAE & AAE & SAE & AAE \\
\midrule
\multirow{3}{*}{Mistral-Small-3.1-24B}
& \textbf{Ours (Best)}  & \textbf{71.01} & \textbf{28.99} & \textbf{56.58} & \textbf{43.42} \\
& Most Likely  & 71.20 & 28.80 & 53.20 & 46.80 \\
& Stylistic    & 73.40 & 26.60 & 50.80 & 49.20 \\
\midrule
\multirow{3}{*}{Phi-4}
& \textbf{Ours (Best)}  & \textbf{79.11} & \textbf{20.89} & \textbf{59.92} & \textbf{40.08} \\
& Most Likely  & 79.70 & 20.30 & 66.70 & 33.30 \\
& Stylistic    & 79.40 & 20.60 & 62.70 & 37.30 \\
\midrule
\multirow{3}{*}{Gemma-3}
& \textbf{Ours (Best)}  & \textbf{66.25} & \textbf{33.75} & \textbf{36.64} & \textbf{63.36} \\
& Most Likely  & 67.80 & 32.20 & 40.46 & 59.54 \\
& Stylistic    & 69.50 & 31.50 & 43.10 & 56.90 \\
\midrule
\multirow{3}{*}{Phi3-Medium}
& \textbf{Ours (Best)}  & \textbf{79.92} & \textbf{20.08} & \textbf{69.61} & \textbf{30.39} \\
& Most Likely  & 78.10 & 21.90 & 66.80 & 32.20 \\
& Stylistic    & 78.80 & 21.20 & 62.70 & 37.30 \\
\midrule
\multirow{3}{*}{DeepSeek-R1}
& \textbf{Ours (Best)}  & \textbf{49.45} & \textbf{50.55} & \textbf{49.50} & \textbf{50.50} \\
& Most Likely  & 50.33 & 49.67 & 49.20 & 50.80 \\
& Stylistic    & 51.62 & 48.38 & 48.87 & 51.13 \\
\midrule
\multirow{3}{*}{Llama-3.1-70B}
& \textbf{Ours (Best)}  & \textbf{61.01} & \textbf{38.99} & \textbf{55.45} & \textbf{44.55} \\
& Most Likely  & 60.53 & 39.47 & 57.09 & 42.91 \\
& Stylistic    & 61.72 & 38.28 & 54.19 & 45.81 \\
\bottomrule
\end{tabular}
}
\caption{Prompt wording ablation for multiple-choice continuation preference.}
\label{tab:prompt_ablation}
\end{table}

\subsection{Layer Selection Ablation}
\label{app:single-vs-multi}

Table~\ref{tab:ablation-steering} asks whether the causal-tracing layer
selection actually matters, or whether steering any reasonable set of
layers would work equally well. We compare four strategies. \emph{Single}
applies the steering direction only at the most causally important layer. \emph{Random-4} steers four
randomly chosen layers and \emph{Middle-4} steers the four middle layers
of the network, both with equal per-layer weights. \emph{Ours} applies
weighted steering across the top-4 traced layers. All strategies share
the same dialect directions. For each strategy, we sweep the steering
strength and report its best setting per model.

Two patterns emerge. First, distributing the intervention over four
layers is usually better than concentrating it in one. Random-4 lowers
LP relative to Single for five of the six models, and Middle-4 for
three. The debiasing signal is therefore not captured by any single
layer. Second, where those four layers sit matters more. Ours reaches
the lowest LP and the lowest AAE perplexity for every model, with a
margin of 2 to 5 LP points over the next-best strategy. SAE perplexity
stays within about one point of the alternatives throughout. Untraced
placements thus recover only part of the effect. The remaining gain
comes specifically from steering the causally traced layers, weighted
by their traced importance, and it arrives at no additional utility
cost.

\begin{table}[h]
\centering
\footnotesize
\setlength{\tabcolsep}{4pt}
\renewcommand{\arraystretch}{1.0}

\resizebox{\columnwidth}{!}{
\begin{tabular}{lccrrr}
\toprule
Model
& Strategy
& $\boldsymbol{\beta}$
& LP
& \makecell{AAE \\ PPL}
& \makecell{SAE \\ PPL} \\
\midrule

\multirow{4}{*}{Phi-4}
& Single        & 0.4 & 50.71 & 131.36 & 27.77 \\
& Random-4      & 0.6  & 52.36    & 133.59     & 28.72    \\
& Middle-4     & 0.6  & 53.08    & 136.32     & 27.66    \\
& \textbf{Ours}         & \textbf{0.6} & \textbf{48.22} & \textbf{107.44} & \textbf{27.92} \\
\midrule

\multirow{4}{*}{Phi-3}
& Single         & 0.6 & 65.73 & 90.53 & 24.89 \\
& Random-4      & 0.6  & 63.51    & 87.25    & 24.94    \\
& Middle-4     & 0.6  & 66.86    & 92.43    & 24.76    \\
& \textbf{Ours}  & \textbf{0.6} & \textbf{60.34} & \textbf{82.83} & \textbf{24.63} \\
\midrule

\multirow{4}{*}{Mistral}
& Single        & 0.6 & 54.73 & 77.78 & 23.71 \\
& Random-4     & 0.8  & 53.75  & 77.62    & 24.47   \\
& Middle-4     & 0.6  & 56.22    & 79.18    & 23.54    \\
& \textbf{Ours} & \textbf{0.8} & \textbf{49.71} & \textbf{77.31} & \textbf{23.93} \\
\midrule

\multirow{4}{*}{DeepSeek-R1}
& Single        & 1.0 & 67.00 & 137.38 & 27.72 \\
& Random-4      & 1.0  & 65.84    & 136.25     & 27.93    \\
& Middle-4      & 1.0  & 65.53    & 136.11     & 27.86    \\
& \textbf{Ours} & \textbf{1.0} & \textbf{60.80} & \textbf{133.08} & \textbf{28.32} \\
\midrule

\multirow{4}{*}{Gemma-3}
& Single       & 1.0 & 88.71 & 132.73 & 56.26 \\
& Random-4     & 1.0  & 86.26    & 131.99       & 55.94      \\
& Middle-4     & 0.8  & 85.34    & 136.25       & 56.52      \\
& \textbf{Ours}& \textbf{0.8} & \textbf{82.91} & \textbf{121.93} & \textbf{56.57} \\
\midrule

\multirow{4}{*}{Llama-3.1-70B}
& Single       & 0.6 & 34.72 & 65.78 & 15.50 \\
& Random-4     & 0.6  & 32.68    & 64.95    & 15.59    \\
& Middle-4     & 0.8  & 31.24    & 64.61    & 15.77    \\
& \textbf{Ours} & \textbf{1.0} & \textbf{27.56} & \textbf{64.08} & \textbf{15.68} \\
\bottomrule
\end{tabular}
}

\caption{Layer selection ablation: steering the single most-traced
layer (Single), four random layers (Random-4), the four middle layers
(Middle-4), or the top-4 traced layers with causal-importance weights
(Ours). For each strategy, we report its best $\beta$ per model. Lower
is better for all three metrics.}
\label{tab:ablation-steering}
\end{table}

\subsection{Full Steering Results}
\label{app:full-steering}

\begin{table*}[!t]
\centering
\scriptsize
\setlength{\tabcolsep}{4pt}
\begin{tabular}{llcccccccccc}
\toprule
Model & $\beta$ & LP $\downarrow$ & Bias Red. $\uparrow$ & AAE PPL $\downarrow$ & AAE PPL Ratio & SAE PPL & SAE PPL Ratio & AAE Match & SAE Match & DGI & cDGI\\
\midrule
\multirow{7}{*}{\textbf{Phi-4}}
& 0.0 & 58.39 & 0.0 & 121.90 & 1.00 & 26.52 & 1.00 & 38.17 & 89.10 & 0.73 & 0.75\\
& 0.1 & 57.88 & 0.9 & 116.10 & 0.95 & 26.57 & 1.00 & 41.59 & 89.92 & 0.74 & 0.76\\
& 0.2 & 53.68 & 8.1 & 112.95 & 0.93 & 27.52 & 1.04 & 44.74 & 89.62 & 0.76 & 0.77\\
& 0.4 & 46.61 & 20.2 & 110.16 & 0.90 & 28.99 & 1.09 & 49.28 & 89.12 & 0.80 & 0.81\\
& \textbf{0.6} & \textbf{47.25} & \textbf{19.1} & \textbf{111.44} & \textbf{0.91} & \textbf{28.72} & \textbf{1.08} & \textbf{58.82} & \textbf{88.25} & \textbf{0.85} & \textbf{0.86}\\
& 0.8 & 46.42 & 20.5 & 136.49 & 1.12 & 41.70 & 1.57 & 66.52 & 75.89 & 0.86 & 0.88\\
& 1.0 & 45.16 & 22.7 & 140.32 & 1.15 & 45.17 & 1.70 & 70.36 & 68.09 & 0.87 & 0.88\\
\midrule
\multirow{7}{*}{\textbf{Phi-3}}
& 0.0 & 71.58 & 0.0 & 89.11 & 1.00 & 20.27 & 1.00 & 33.77 & 79.23 & 0.73 & 0.74\\
& 0.1 & 70.05 & 2.1 & 84.62 & 0.95 & 20.57 & 1.01 & 37.40 & 79.88 & 0.73 & 0.73\\
& 0.2 & 69.01 & 3.6 & 83.42 & 0.94 & 20.57 & 1.01 & 40.85 & 78.96 & 0.73 & 0.74\\
& 0.4 & 65.92 & 7.9 & 81.53 & 0.91 & 20.72 & 1.02 & 48.34 & 77.45 & 0.74 & 0.75\\
& \textbf{0.6} & \textbf{61.53} & \textbf{14.0} & \textbf{80.53} & \textbf{0.90} & \textbf{25.03} & \textbf{1.23} & \textbf{55.26} & \textbf{77.34} & \textbf{0.77} & \textbf{0.78}\\
& 0.8 & 62.10 & 13.2 & 106.10 & 1.19 & 30.83 & 1.52 & 58.10 & 71.83 & 0.77 & 0.78\\
& 1.0 & 62.07 & 13.3 & 117.76 & 1.32 & 34.53 & 1.70 & 62.82 & 66.29 & 0.78 & 0.78\\
\midrule
\multirow{7}{*}{\textbf{Mistral}}
& 0.0 & 59.93 & 0.0 & 86.91 & 1.00 & 23.57 & 1.00 & 28.04 & 89.42 & 0.78 & 0.80\\
& 0.1 & 56.85 & 5.1 & 81.92 & 0.94 & 23.28 & 0.99 & 31.32 & 89.30 & 0.79 & 0.80\\
& 0.2 & 56.45 & 5.8 & 81.08 & 0.93 & 23.26 & 0.99 & 35.91 & 88.11 & 0.81 & 0.82\\
& 0.4 & 53.60 & 10.6 & 77.48 & 0.89 & 23.25 & 0.99 & 41.85 & 88.08 & 0.83 & 0.84\\
& 0.6 & 51.73 & 13.7 & 76.78 & 0.88 & 23.18 & 0.98 & 46.27 & 87.67 & 0.84 & 0.85\\
& \textbf{0.8} & \textbf{50.21} & \textbf{16.2} & \textbf{75.91} & \textbf{0.87} & \textbf{24.13} & \textbf{1.02} & \textbf{48.18} & \textbf{88.63} & \textbf{0.85} & \textbf{0.86}\\
& 1.0 & 50.01 & 16.6 & 75.99 & 0.87 & 24.88 & 1.06 & 49.73 & 85.87 & 0.85 & 0.86\\
\midrule
\multirow{7}{*}{\textbf{DeepSeek-R1}}
& 0.0 & 71.98 & 0.0 & 145.84 & 1.00 & 27.78 & 1.00 & 42.13 & 76.15 & 0.56 & 0.58\\
& 0.1 & 67.73 & 5.9 & 141.01 & 0.97 & 28.10 & 1.01 & 47.92 & 76.06 & 0.56 & 0.57\\
& 0.2 & 66.83 & 7.2 & 139.41 & 0.96 & 28.08 & 1.01 & 51.27 & 76.39 & 0.58 & 0.58\\
& 0.4 & 64.19 & 10.8 & 137.40 & 0.94 & 28.02 & 1.01 & 56.58 & 74.74 & 0.59 & 0.59\\
& 0.6 & 63.91 & 11.2 & 136.57 & 0.94 & 27.98 & 1.01 & 60.46 & 73.61 & 0.60 & 0.62\\
& 0.8 & 62.51 & 13.2 & 136.49 & 0.94 & 27.96 & 1.01 & 62.72 & 73.21 & 0.61 & 0.62\\
& \textbf{1.0} & \textbf{62.00} & \textbf{13.9} & \textbf{138.38} & \textbf{0.95} & \textbf{27.92} & \textbf{1.01} & \textbf{64.43} & \textbf{71.07} & \textbf{0.62} & \textbf{0.63}\\
\midrule
\multirow{7}{*}{\textbf{Gemma-3}}
& 0.0 & 93.07 & 0.0 & 138.82 & 1.00 & 56.42 & 1.00 & 44.39 & 84.10 & 0.79 & 0.81\\
& 0.1 & 87.40 & 6.1 & 138.09 & 0.99 & 56.55 & 1.00 & 44.33 & 81.03 & 0.80 & 0.81\\
& 0.2 & 87.03 & 6.5 & 136.39 & 0.98 & 56.93 & 1.01 & 49.58 & 83.57 & 0.81 & 0.81\\
& 0.4 & 85.92 & 7.7 & 133.01 & 0.96 & 57.03 & 1.01 & 54.17 & 82.00 & 0.82 & 0.82\\
& 0.6 & 83.78 & 10.0 & 131.87 & 0.95 & 57.36 & 1.02 & 57.84 & 82.93 & 0.83 & 0.84\\
& \textbf{0.8} & \textbf{83.01} & \textbf{10.8} & \textbf{124.58} & \textbf{0.90} & \textbf{57.40} & \textbf{1.02} & \textbf{60.78} & \textbf{83.19} & \textbf{0.86} & \textbf{0.87}\\
& 1.0 & 83.71 & 10.1 & 123.50 & 0.89 & 58.28 & 1.03 & 62.49 & 81.71 & 0.86 & 0.86\\
\midrule
\multirow{7}{*}{\textbf{Llama-3.1-70B}}
& 0.0 & 34.46 & 0.0 & 68.93 & 1.00 & 15.49 & 1.00 & 42.95 & 89.35 & 0.52 & 0.57\\
& 0.1 & 33.83 & 1.8 & 65.36 & 0.95 & 15.48 & 1.00 & 42.62 & 88.32 & 0.53 & 0.55\\
& 0.2 & 33.79 & 1.9 & 64.68 & 0.94 & 15.48 & 1.00 & 45.41 & 87.97 & 0.54 & 0.56\\
& 0.4 & 31.26 & 9.3 & 63.33 & 0.92 & 15.49 & 1.00 & 54.18 & 88.04 & 0.57 & 0.60\\
& 0.6 & 30.72 & 10.9 & 63.09 & 0.92 & 15.50 & 1.00 & 60.30 & 87.51 & 0.59 & 0.61\\
& 0.8 & 30.51 & 11.5 & 62.11 & 0.90 & 15.47 & 1.00 & 62.45 & 87.89 & 0.62 & 0.63\\
& \textbf{1.0} & \textbf{29.76} & \textbf{13.6} & \textbf{62.18} & \textbf{0.90} & \textbf{15.48} & \textbf{1.00} & \textbf{64.75} & \textbf{85.99} & \textbf{0.63} & \textbf{0.64}\\
\bottomrule
\end{tabular}
\caption{Full steering results across all tested $\beta$ values computed on the validation set. Bias reduction and perplexity ratios are computed relative to $\beta = 0$ for each model. Bold marks the selected best $\beta$ value for each model, chosen to balance lower bias against preservation of SAE fluency (SAE PPL and SAE Match).}
\label{tab:debiasing-appendix-full}
\end{table*}

Table~\ref{tab:debiasing-appendix-full} reports the full multi-layer steering sweep, and the trends across $\beta$. In the table, \textbf{LP} measures log-probability bias, where lower values indicate less preference for SAE continuations. \textbf{AAE Match} and \textbf{SAE Match} are multiple-choice consistency scores under AAE and SAE contexts, respectively, where higher values are better. \textbf{AAE PPL} and \textbf{SAE PPL} are the perplexities of the AAE-matched and SAE-matched continuations, respectively, where lower values indicate better support for the matched continuation. \textbf{DGI} and \textbf{cDGI} measure prediction consistency across dialectal variants, with higher values indicating better invariance.

\noindent\textbf{(a) Discriminative consistency (DGI, cDGI):} Both metrics increase monotonically with $\beta$ for all models (e.g., Phi-4: DGI $0.74 \rightarrow 0.87$, cDGI $0.74 \rightarrow 0.88$; Llama-3.1-70B: DGI $0.51 \rightarrow 0.63$, cDGI $0.54 \rightarrow 0.64$), showing that steering also improves AAE--SAE prediction agreement on the discriminative task, not just continuation preference.

\noindent\textbf{(b) Perplexity (PPL).} AAE PPL decreases at small to moderate $\beta$ (e.g., Phi-4: $121.90 \rightarrow 110.16$ at $\beta = 0.4$; Mistral: $86.91 \rightarrow 75.91$ at $\beta = 0.8$), while SAE PPL stays within 2 points of $\beta = 0$ for Mistral, DeepSeek-R1, Gemma-3, and Llama-3.1-70B at every $\beta$. For Phi-4 and Phi-3, AAE PPL rebounds at $\beta \geq 0.8$ and SAE PPL rises sharply (Phi-4: $26.52 \rightarrow 45.17$; Phi-3: $20.27 \rightarrow 34.53$ at $\beta = 1.0$), the same over-steering effect seen in SAE Match.

\noindent\textbf{(c) Log-probability bias (LP).} LP decreases overall with $\beta$ for all six models, from $58.39$ to $45.16$ on Phi-4, $71.58$ to $62.07$ on Phi-3, and $34.46$ to $29.76$ on Llama-3.1-70B, a relative reduction of 10\% to 23\% at $\beta = 1.0$. Most of the reduction is reached by $\beta = 0.6$, and LP changes little beyond $\beta = 0.8$.

\noindent\textbf{(d) Multiple-choice preference (MC).} AAE Match is higher than at $\beta = 0$ for every $\beta \geq 0.2$ and peaks at $\beta = 1.0$ for every model (e.g., Phi-4: $38.17 \rightarrow 70.36$; DeepSeek-R1: $42.13 \rightarrow 64.43$; Llama-3.1-70B: $42.95 \rightarrow 64.75$). SAE Match stays within about 5 points of $\beta = 0$ for Mistral, DeepSeek-R1, Gemma-3, and Llama-3.1-70B, but drops at high $\beta$ for Phi-4 ($89.10 \rightarrow 68.09$) and Phi-3 ($79.23 \rightarrow 66.29$), so over-steering erodes SAE behavior for these two models.

Across models, most of the LP and MC gains are reached by $\beta \in [0.4, 0.8]$ while SAE PPL and SAE Match stay close to $\beta = 0$, and the selected values (bold in Table~\ref{tab:debiasing-appendix-full}) fall in this range for four models. For DeepSeek-R1 and Llama-3.1-70B, SAE PPL and SAE Match stay flat up to $\beta = 1.0$, so the largest value is selected. For the Phi models, $\beta = 1.0$ adds little LP reduction but hurts SAE fluency.

\subsection{Qualitative Example of a Cross-Model Preference Flip}
\label{app:qualitative-steering}

\paragraph{Base-model preference under AAE context.}
Table~\ref{tab:pmc-example} shows two representative cases in
which Mistral prefers an SAE continuation under an AAE context.

\paragraph{Steering-induced preference flip.}
Table~\ref{tab:qualitative-steering} shows an example in which both
the full sentence and the evaluated AAE context contain AAE markers,
together with the segmented context--continuation pair used in
evaluation. Before steering, five of the six models prefer the SAE
continuation under this AAE context; after steering, five of the six
prefer the AAE continuation.

\begin{table}[h]
\footnotesize
\centering
\begin{tabular}{p{0.09\textwidth} p{0.33\textwidth}}
\toprule
\textbf{AAE ctx.} & ``Let's see what my tl have to ...'' \\
\textbf{SAE cont.} & ``... offer - nonsense, nonsense, and more nonsense.'' \textit{(Model Preferred)}\\
\textbf{AAE cont.} & ``... offer bull ish bull ish n more bull ish.'' \\
\addlinespace[4pt]
\midrule
\addlinespace[4pt]
\textbf{AAE ctx.} & ``My Fone Kno Me So Well ...'' \\
\textbf{SAE cont.} & ``... when I'm on hold for less than one minute, it will hang up.'' \textit{(Model Preferred)}\\
\textbf{AAE cont.} & ``... When Im On Hold For less Then 1 Minute it Would Hang Upp.'' \\
\bottomrule
\end{tabular}
\caption{Mistral-Small-3.1-24B prefers the SAE
\textit{continuation} (cont.) under AAE \textit{context}
(ctx.), effectively ``correcting'' the dialect.}
\label{tab:pmc-example}
\end{table}

\begin{table*}[t]
\centering
\small
\setlength{\tabcolsep}{5pt}

\begin{tabular}{p{0.18\textwidth}p{0.75\textwidth}}
\toprule
Full AAE sentence &
\textit{That is why I wanna go back so badly. I wanna disprove any negatives I brought to me and my family's name.} \\
AAE context &
\textit{That is why I wanna go} \\
AAE continuation &
\textit{back so badly. I wanna disprove any negatives I brought to me and my family's name.} \\
SAE continuation &
\textit{back so badly. I want to disprove any negatives I brought to myself and my family's name.} \\
\midrule
Cross-model vote &
Before debiasing, \textbf{5/6} models prefer the SAE continuation under this AAE
context; after debiasing, \textbf{5/6} models prefer the AAE continuation. \\
\bottomrule
\end{tabular}
\caption{A qualitative example of steering-induced preference reversal. The full
source sentence is recognizably AAE, and the evaluated AAE context itself contains
AAE markers.}
\label{tab:qualitative-steering}
\end{table*}

\subsection{Full Mitigation Results with Confidence Intervals}
\label{app:table4_ci}

Table~\ref{tab:mitigation_ci} reports all mitigation methods,
including the zero- and few-shot prompting baselines, with 95\%
paired bootstrap confidence intervals over the 2{,}000 test pairs.
For each bootstrap sample, we resample evaluation pairs with
replacement and recompute all metrics for every method on the same
resampled set, which preserves the paired structure across methods.

Zero-shot and few-shot prompting raise AAE Match by 2 to 6 and 5 to
10 points, respectively, compared with 18 to 24 points for
steering, and steering reaches the best mean on every bias metric
for all six models.

\begin{table*}[!t]
\centering
\scriptsize
\setlength{\tabcolsep}{3.2pt}
\renewcommand{\arraystretch}{1.05}
\newcommand{\ours}[1]{\cellcolor{green!10}#1}
\begin{tabular}{llcccccc}
\toprule
\textbf{Metric} & \textbf{Method} & \textbf{Phi-4} & \textbf{Phi-3} & \textbf{Mistral} &
\textbf{DeepSeek-R1} & \textbf{Gemma-3} & \textbf{Llama-3.1-70B} \\
\midrule
DGI $\uparrow$ & Base & $0.74 \pm 0.02$ & $0.73 \pm 0.02$ & $0.79 \pm 0.02$ & $0.55 \pm 0.01$ & $0.80 \pm 0.02$ & $0.51 \pm 0.01$ \\
 & Zero-shot & $0.76 \pm 0.02$ & $0.74 \pm 0.02$ & $0.83 \pm 0.02$ & $0.57 \pm 0.01$ & $0.80 \pm 0.02$ & $0.55 \pm 0.01$ \\
 & Few-shot & $0.79 \pm 0.02$ & $0.74 \pm 0.02$ & $0.84 \pm 0.02$ & $0.58 \pm 0.01$ & $0.81 \pm 0.02$ & $0.61 \pm 0.01$ \\
 & CAD & $0.76 \pm 0.02$ & $0.75 \pm 0.02$ & $0.83 \pm 0.02$ & $0.56 \pm 0.01$ & $0.81 \pm 0.02$ & $0.53 \pm 0.01$ \\
 & LoRA & $0.78 \pm 0.02$ & $0.75 \pm 0.02$ & $0.83 \pm 0.02$ & $0.57 \pm 0.01$ & $0.82 \pm 0.02$ & $0.56 \pm 0.01$ \\
 & \ours{Ours} & \ours{$\mathbf{0.86} \pm 0.02$} & \ours{$\mathbf{0.77} \pm 0.02$} & \ours{$\mathbf{0.86} \pm 0.02$} & \ours{$\mathbf{0.63} \pm 0.01$} & \ours{$\mathbf{0.85} \pm 0.02$} & \ours{$\mathbf{0.65} \pm 0.01$} \\
\midrule
cDGI $\uparrow$ & Base & $0.77 \pm 0.02$ & $0.74 \pm 0.02$ & $0.81 \pm 0.02$ & $0.59 \pm 0.01$ & $0.82 \pm 0.02$ & $0.57 \pm 0.01$ \\
 & Zero-shot & $0.79 \pm 0.02$ & $0.75 \pm 0.02$ & $0.84 \pm 0.02$ & $0.60 \pm 0.01$ & $0.83 \pm 0.02$ & $0.59 \pm 0.01$ \\
 & Few-shot & $0.80 \pm 0.02$ & $0.75 \pm 0.02$ & $0.85 \pm 0.02$ & $0.60 \pm 0.01$ & $0.82 \pm 0.02$ & $0.62 \pm 0.01$ \\
 & CAD & $0.77 \pm 0.02$ & $0.76 \pm 0.01$ & $0.84 \pm 0.02$ & $0.59 \pm 0.01$ & $0.82 \pm 0.02$ & $0.58 \pm 0.01$ \\
 & LoRA & $0.79 \pm 0.02$ & $0.76 \pm 0.02$ & $0.84 \pm 0.02$ & $0.59 \pm 0.01$ & $0.83 \pm 0.02$ & $0.59 \pm 0.01$ \\
 & \ours{Ours} & \ours{$\mathbf{0.87} \pm 0.02$} & \ours{$\mathbf{0.78} \pm 0.02$} & \ours{$\mathbf{0.87} \pm 0.02$} & \ours{$\mathbf{0.64} \pm 0.01$} & \ours{$\mathbf{0.86} \pm 0.02$} & \ours{$\mathbf{0.66} \pm 0.01$} \\
\midrule
AAE PPL $\downarrow$ & Base & $123.9 \pm 8.7$ & $85.8 \pm 6.0$ & $82.8 \pm 5.8$ & $142.8 \pm 10.0$ & $133.4 \pm 9.4$ & $67.7 \pm 4.8$ \\
 & CAD & $124.8 \pm 8.0$ & $86.1 \pm 5.4$ & $84.8 \pm 6.3$ & $143.4 \pm 8.7$ & $135.6 \pm 9.9$ & $69.1 \pm 4.5$ \\
 & LoRA & $113.0 \pm 8.9$ & $83.7 \pm 6.2$ & $78.1 \pm 5.9$ & $140.0 \pm 8.8$ & $129.3 \pm 9.5$ & $64.9 \pm 4.3$ \\
 & \ours{Ours} & \ours{$\mathbf{107.4} \pm 7.8$} & \ours{$\mathbf{82.8} \pm 5.7$} & \ours{$\mathbf{77.3} \pm 5.3$} & \ours{$\mathbf{133.0} \pm 9.6$} & \ours{$\mathbf{121.9} \pm 8.5$} & \ours{$\mathbf{64.1} \pm 4.4$} \\
\midrule
SAE PPL & Base & $26.7 \pm 1.9$ & $20.6 \pm 1.5$ & $23.3 \pm 1.7$ & $28.1 \pm 2.0$ & $56.3 \pm 3.8$ & $15.5 \pm 1.1$ \\
 & CAD & $27.8 \pm 2.1$ & $23.5 \pm 1.5$ & $24.8 \pm 1.6$ & $30.5 \pm 2.3$ & $57.7 \pm 3.9$ & $15.8 \pm 1.0$ \\
 & LoRA & $27.4 \pm 2.1$ & $23.1 \pm 1.4$ & $23.7 \pm 1.8$ & $28.8 \pm 1.9$ & $56.1 \pm 4.1$ & $15.6 \pm 1.0$ \\
 & \ours{Ours} & \ours{$27.9 \pm 2.0$} & \ours{$24.6 \pm 1.8$} & \ours{$23.9 \pm 1.7$} & \ours{$28.3 \pm 2.0$} & \ours{$56.6 \pm 4.0$} & \ours{$15.7 \pm 1.1$} \\
\midrule
AAE Match $\uparrow$ & Base & $39.3 \pm 2.1$ & $32.6 \pm 2.1$ & $29.5 \pm 2.0$ & $43.8 \pm 2.2$ & $42.9 \pm 2.2$ & $41.3 \pm 2.2$ \\
 & Zero-shot & $41.5 \pm 2.1$ & $38.6 \pm 2.2$ & $33.9 \pm 1.9$ & $48.7 \pm 1.9$ & $44.4 \pm 2.2$ & $44.5 \pm 2.2$ \\
 & Few-shot & $44.8 \pm 2.2$ & $42.5 \pm 2.1$ & $36.3 \pm 2.1$ & $51.6 \pm 2.2$ & $48.4 \pm 2.0$ & $48.1 \pm 1.9$ \\
 & CAD & $43.8 \pm 3.1$ & $39.4 \pm 2.5$ & $33.5 \pm 2.5$ & $50.2 \pm 3.7$ & $46.2 \pm 3.1$ & $47.2 \pm 3.5$ \\
 & LoRA & $45.1 \pm 2.7$ & $42.7 \pm 2.8$ & $41.6 \pm 3.2$ & $50.9 \pm 4.0$ & $51.8 \pm 4.0$ & $53.7 \pm 4.1$ \\
 & \ours{Ours} & \ours{$\mathbf{58.3} \pm 2.1$} & \ours{$\mathbf{56.9} \pm 2.2$} & \ours{$\mathbf{50.1} \pm 2.0$} & \ours{$\mathbf{66.2} \pm 1.7$} & \ours{$\mathbf{61.0} \pm 2.0$} & \ours{$\mathbf{65.4} \pm 2.1$} \\
\midrule
SAE Match & Base & $89.1 \pm 1.4$ & $79.7 \pm 1.8$ & $88.6 \pm 1.4$ & $75.5 \pm 1.9$ & $82.9 \pm 1.6$ & $88.6 \pm 1.4$ \\
 & Zero-shot & $90.1 \pm 1.3$ & $82.8 \pm 1.7$ & $89.5 \pm 1.3$ & $71.3 \pm 2.0$ & $83.7 \pm 1.6$ & $86.8 \pm 1.7$ \\
 & Few-shot & $90.8 \pm 1.3$ & $83.1 \pm 1.6$ & $89.2 \pm 1.4$ & $73.7 \pm 1.9$ & $84.9 \pm 1.6$ & $88.1 \pm 1.4$ \\
 & CAD & $89.4 \pm 1.4$ & $79.6 \pm 1.6$ & $88.9 \pm 1.4$ & $73.8 \pm 2.0$ & $83.3 \pm 1.7$ & $88.5 \pm 1.6$ \\
 & LoRA & $88.7 \pm 1.4$ & $78.6 \pm 1.7$ & $88.7 \pm 1.5$ & $73.5 \pm 2.0$ & $83.1 \pm 1.5$ & $87.8 \pm 1.3$ \\
 & \ours{Ours} & \ours{$88.7 \pm 1.4$} & \ours{$78.2 \pm 1.8$} & \ours{$88.4 \pm 1.4$} & \ours{$72.6 \pm 1.9$} & \ours{$83.3 \pm 1.6$} & \ours{$87.7 \pm 1.4$} \\
\midrule
LP $\downarrow$ & Base & $60.6 \pm 4.4$ & $70.8 \pm 5.1$ & $57.3 \pm 4.2$ & $68.9 \pm 5.4$ & $89.3 \pm 6.5$ & $34.9 \pm 2.6$ \\
 & Zero-shot & $58.6 \pm 4.3$ & $69.0 \pm 5.0$ & $56.9 \pm 4.1$ & $68.2 \pm 5.3$ & $88.4 \pm 6.4$ & $35.0 \pm 2.6$ \\
 & Few-shot & $54.5 \pm 4.0$ & $62.1 \pm 4.6$ & $54.1 \pm 3.9$ & $62.7 \pm 4.7$ & $88.1 \pm 6.1$ & $31.9 \pm 2.4$ \\
 & CAD & $61.7 \pm 4.5$ & $72.4 \pm 5.7$ & $58.1 \pm 3.8$ & $70.5 \pm 5.1$ & $92.1 \pm 6.5$ & $37.2 \pm 2.4$ \\
 & LoRA & $55.1 \pm 4.2$ & $64.9 \pm 3.9$ & $53.0 \pm 3.4$ & $66.3 \pm 5.1$ & $86.7 \pm 5.5$ & $31.6 \pm 2.5$ \\
 & \ours{Ours} & \ours{$\mathbf{48.2} \pm 3.5$} & \ours{$\mathbf{60.3} \pm 4.5$} & \ours{$\mathbf{49.7} \pm 3.7$} & \ours{$\mathbf{60.8} \pm 4.6$} & \ours{$\mathbf{82.9} \pm 6.1$} & \ours{$\mathbf{27.6} \pm 2.2$} \\
\bottomrule
\end{tabular}
\caption{All mitigation results on the held-out test split, with 95\%
paired bootstrap confidence intervals over the 2{,}000 test pairs
(mean $\pm$ half-width). \textit{Zero-shot} and \textit{Few-shot}:
prompting baselines (Appendix~\ref{app:prompt_baseline});
\textit{CAD}, \textit{LoRA}, and \textit{Ours} as in
Table~\ref{tab:main-debiasing-results}. \textbf{Bold}: best mean per
model; SAE PPL and SAE Match are not bolded. PPL is not reported for
prompting because it changes the input, so likelihoods are not
comparable.}
\label{tab:mitigation_ci}
\end{table*}

\subsection{cDGI Sensitivity Analysis}
\label{app:cdgi-sensitivity}

cDGI is computed on a filtered subset, so its value may depend on
the conditioning criteria. We consider three settings:
prediction stability plus majority-annotator naturalness (our
default), prediction stability plus unanimous naturalness, and
prediction stability alone. For each naturalness criterion, the
annotator-natural pool is fixed across models, whereas prediction
stability is computed separately from each model's outputs.
Consequently, the final cDGI conditioning set is model-specific.

Table~\ref{tab:cdgi-sensitivity} reports cDGI under each setting.
The six-model ordering is unchanged, indicating that the model
ranking is robust to these alternative conditioning criteria.

\begin{table}[t]
\centering
\footnotesize
\setlength{\tabcolsep}{4pt}
\renewcommand{\arraystretch}{1.05}
\resizebox{\columnwidth}{!}{
\begin{tabular}{lcccccc}
\toprule
Setting & Gemma-3 & Phi-4 & Mistral & Phi-3 & \makecell{DeepSeek\\-R1} & \makecell{Llama\\-70B} \\
\midrule
majority (default) & .866 & .827 & .815 & .792 & .583 & .525 \\
unanimous           & .860 & .816 & .798 & .787 & .595 & .485 \\
Stability only               & .889 & .835 & .824 & .786 & .660 & .561 \\
\bottomrule
\end{tabular}
}
\caption{cDGI under alternative conditioning criteria. The
six-model ordering is unchanged across settings. The majority-
and unanimous-naturalness pools are fixed across models, whereas
the prediction-stability filter and resulting cDGI conditioning
set are model-specific. ``Stability only'' omits the naturalness
filter.}
\label{tab:cdgi-sensitivity}
\end{table}

\subsection{Bias--Utility trade-off}
\label{app:tradeoff}
\begin{figure*}[!t]
    \centering
    \includegraphics[width=\textwidth]{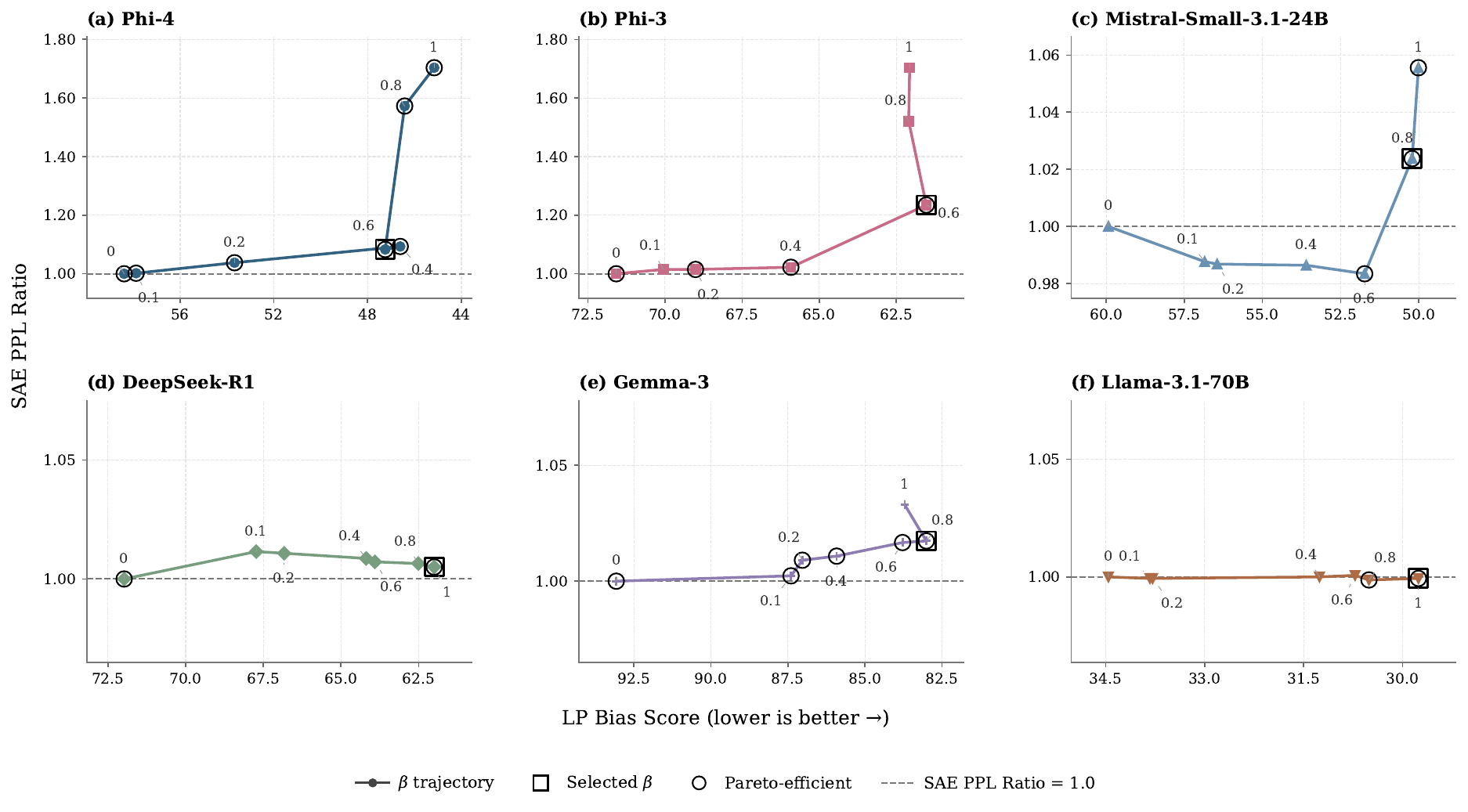}
    \caption{Bias--utility frontier across steering strengths $\beta$ for all six models. Each point corresponds to one $\beta$ setting. The x-axis shows LP bias score (lower is better), and the y-axis shows SAE perplexity ratio relative to $\beta=0$ (values at or below the dashed line indicate no loss in SAE fluency). }
    \label{fig:beta_frontier}
\end{figure*}

In Fig.~\ref{fig:beta_frontier}, we show how the bias and utility change by sweeping over $\beta$.
  Each point corresponds to one $\beta$ setting
for one model. The x-axis shows the LP bias score, where lower values indicate less
preference for SAE continuations, and the y-axis shows the SAE perplexity ratio
relative to the unsteered baseline $\beta=0$, where values close to 1.0 indicate
better preservation of SAE fluency.

\subsection{Capability Preservation under Steering}
\label{app:capability}

A standing activation intervention could in principle degrade
abilities unrelated to dialect. We test this by evaluating each model on
MMLU \citep{hendrycks2021measuring} (a 500-question stratified
subset, fixed seed, 5-shot) and GSM8K \citep{cobbe2021gsm8k}
(200 questions, 8-shot) with and without steering at the
selected $\beta$. Both conditions use identical
prompts.

Table~\ref{tab:capability} reports accuracy. Deltas range from
$-1.5$ to $+2.5$ points, well
within the sampling error of the evaluation subsets (about
$\pm 3.5$ points for MMLU and $\pm 3.8$ points for GSM8K at 95\%
confidence). We find no evidence of capability degradation from steering at the selected strength.

\begin{table}[!htbp]
\centering
\footnotesize
\setlength{\tabcolsep}{4pt}
\renewcommand{\arraystretch}{1.05}
\resizebox{\columnwidth}{!}{
\begin{tabular}{lcccccc}
\toprule
& \multicolumn{3}{c}{MMLU} & \multicolumn{3}{c}{GSM8K} \\
\cmidrule(lr){2-4}\cmidrule(lr){5-7}
Model & Base & Steered & $\Delta$ & Base & Steered & $\Delta$ \\
\midrule
Phi-4         & 80.0 & 81.0 & $+1.0$ & 92.5 & 91.0 & $-1.5$ \\
Phi-3         & 76.2 & 76.8 & $+0.6$ & 89.0 & 89.5 & $+0.5$ \\
Mistral       & 82.0 & 82.4   & $+0.4$    & 93.0 & 93.5   & $+0.5$   \\
DeepSeek-R1   & 81.4 & 81.4   &   $+0.0$   & 90.0   & 91.0   & $+1.0$     \\
Gemma-3       & 79.6 & 79.8  & $+0.2$     & 91.0   & 91.5   & $+0.5$   \\
Llama-3.1-70B & 80.2 & 81.4 & $+1.2$ & 93.0 & 95.5 & $+2.5$ \\
\bottomrule
\end{tabular}
}
\caption{General-capability benchmarks with and without steering
at the selected per-model $\beta$. All deltas lie within the
sampling error of the evaluation subsets.}
\label{tab:capability}
\end{table}

\subsection{Human Evaluation of Generated Continuations}
\label{app:generation}

\paragraph{Items.} We select 200 posts from \textsc{Real-AAE} that are
not used for layer selection, direction extraction, or the choice
of $\beta$. Each AAE context contains at least one grammatical AAE
marker, and the items are stratified by the main marker: negative
concord (43), future markers such as \emph{finna} and \emph{gon}
(40), copula absence (39), \emph{ain't} (30), habitual \emph{be}
(18), and other markers (30). We exclude posts with slurs or
explicit sexual terms, since models often refuse or rewrite them.

\paragraph{Generation.} For each AAE context, we used the
continuations with and without steering, at the $\beta$ selected in
\S\ref{sec:setup} and with the same decoding settings.

\paragraph{Annotation.} The two continuations of each context are
shown in random order, without the condition. For each pair,
annotators choose which continuation better matches the dialect of
the context and which reads more naturally, with a tie option for
both, and flag any continuation whose AAE reads as exaggerated. All
three annotators label all 200 pairs. We report the share of
judgments averaged over the three annotators, and agreement as
Gwet's AC1 over the 200 pairs.

\paragraph{Results.} Table~\ref{tab:generation-human} shows that
annotators preferred the steered continuation for dialect match in
73.0\% of judgments and the base continuation in 9.8\%, with 17.2\%
ties. The steered continuation was also judged more natural (63.3\%
vs.\ 10.6\%). It was flagged as exaggerated in 5.1\% of pairs,
compared with 35.2\% for the base continuation, so steering does not
make the model overuse AAE features. AC1 is 0.66 for dialect match,
0.75 for naturalness, and 0.71 for exaggeration.

\begin{table}[h]
\centering
\small
\begin{tabular}{lccc}
\toprule
\textbf{Criterion} & \textbf{Steered} & \textbf{Base} & \textbf{Tie} \\
\midrule
Dialect match & 73.0\% & 9.8\% & 17.2\% \\
Naturalness   & 63.3\% & 10.6\% & 26.1\% \\
\midrule
Flagged as exaggerated & 5.1\% & 35.2\% & -- \\
\midrule
\multicolumn{4}{l}{Agreement over 200 pairs (AC1): dialect 0.66,} \\
\multicolumn{4}{l}{naturalness 0.75, exaggerated 0.71} \\
\bottomrule
\end{tabular}
\caption{Pairwise human judgments of base and steered continuations
for 200 AAE contexts (\% of pairs in which each continuation is
preferred, averaged over annotators for shared pairs).}
\label{tab:generation-human}
\end{table}

\section{Annotation Guidelines}
\label{app:annotation-guidelines}

Both annotation tasks used a browser-based interface that saved
answers locally; annotators exported their answers as a file when
they finished. The questions and options below are reproduced
verbatim from the interfaces.

\subsection{Corpus Validation}
\label{app:guidelines-validation}

For the human validation in \S\ref{construction}, each item shows
the original AAE tweet, its SAE translation, and the back-translated
AAE. Before the first item, annotators answered a one-time profile
(whether they identify as AAE speakers, the regional background of
the AAE they know best, their familiarity with written and
social-media AAE, and how they wish to be acknowledged). We report
Q1, Q2, Q3, and Q5 in \S\ref{construction}.

\begin{promptbox}[Corpus validation: instructions and questions]
\small
\textit{Read the original AAE, the SAE translation, and the
back-translated AAE, then answer all required questions.}\\[4pt]
\textbf{Q1.} Does the SAE version convey the same meaning as the
original AAE tweet?\\
{[}Same meaning{]} {[}Partly the same meaning{]} {[}Meaning changed{]}\\
Optional: If not fully, what was lost or changed?\\[3pt]
\textbf{Q2.} Does the original tweet sound like natural AAE -- something
you or someone you know might actually say or write?\\
{[}Natural AAE{]} {[}Somewhat natural / mixed{]} {[}Unnatural or too
noisy to tell{]}\\[3pt]
\textbf{Q3.} Does this back-translated version sound like natural AAE?\\
{[}Natural AAE{]} {[}Somewhat natural / mixed{]} {[}Unnatural or sounds
imitated{]}\\[3pt]
\textbf{Q4.} What is the overall sentiment of the original AAE tweet?\\
{[}Positive{]} {[}Neutral{]} {[}Negative{]}\\[3pt]
\textbf{Q5.} Does the SAE translation preserve the sentiment of the
original AAE tweet?\\
{[}Yes{]} {[}No{]}\\[3pt]
\textbf{Q6.} Does the original AAE tweet contain any word or phrase that
may look toxic or offensive on the surface, even if the overall
meaning is not toxic?\\
{[}No potentially toxic wording{]} {[}Potentially toxic wording, but
non-toxic meaning{]} {[}Toxic wording with toxic or hostile
meaning{]}\\[3pt]
\textbf{Optional flags.} Mark this example for special follow-up or
review. $\cdot$ The original AAE tweet cannot be reliably understood as
written (for example, missing context).
\end{promptbox}

\subsection{Comparison of Generated Continuations}
\label{app:guidelines-generation}

For the evaluation in Appendix~\ref{app:generation}, each item shows
the beginning of a post and two continuations, A and B. The
position of the steered continuation (A or B) is randomized. Annotators first read the guidelines below,
which stay available during the task.

\begin{promptbox}[Generation comparison: guidelines]
\small
\textbf{Content warning:} the posts are real social-media posts. Some
contain profanity, insults or sexual references.\\[3pt]
Each item shows the beginning of a post (the context), written in
African American English (AAE), and two continuations, A and B,
written by computer programs. Read the context, read both
continuations, then answer three questions. A and B are in random
order; there is no pattern to learn. Please judge the words
themselves. Ignore capitalization and punctuation (e.g.\ ``I'' vs
``i'', a missing period).\\[4pt]
\textbf{Q1.} Which continuation better keeps the way the original
writer talks (their dialect and style)?\\
Think of the writer's grammar (e.g.\ \emph{ain't}, habitual
\emph{be}, \emph{finna}, double negatives, dropping \emph{is/are}),
word choice and spelling. Pick the one that sounds more like the
same person kept writing. A continuation that switches into textbook
Standard English keeps the style less well. This is not about which
one you like better or which is ``more correct''.\\
\textit{Example.} Context: she be workin late every day.
A: \ldots so she ain't never home for dinner.
B: \ldots so she is never home for dinner.
$\rightarrow$ A keeps the writer's way of talking better.\\
{[}A{]} {[}B{]} {[}About the same{]}\\[3pt]
\textbf{Q2.} Which continuation follows more naturally from the
context?\\
Does it make sense as the next part of this post: on topic, no
contradiction, not just repeating the context, not garbled? Judge
meaning and flow here, not dialect.\\
{[}A{]} {[}B{]} {[}About the same{]}\\[3pt]
\textbf{Q3.} Does either continuation use AAE in a way that sounds
exaggerated, forced, or like a stereotype?\\
For example: piling up dialect features in a way a real speaker
would not, using features incorrectly, or reading like an imitation
or mockery of how people talk. Tick A, B, both, or Neither.\\
{[}A{]} {[}B{]} {[}Neither{]}\\[3pt]
Choose \emph{About the same} when you honestly see no difference.
If an item cannot be judged (e.g.\ you cannot understand it, or you
prefer not to rate it), tick \emph{Cannot judge}; a short note
helps.
\end{promptbox}

\raggedbottom

\end{document}